\documentclass[10pt,twocolumn,letterpaper]{article}

\usepackage{wacv}
\usepackage{times}
\usepackage{epsfig}
\usepackage{graphicx}
\usepackage{amsmath}
\usepackage{amssymb}
\usepackage{booktabs}
\usepackage{multirow,tabularx}

\usepackage[utf8]{inputenc}
\usepackage[table]{xcolor}
\usepackage{enumitem}
\usepackage[nocompress]{cite}
\usepackage{subcaption}
\newcommand{\ssymbol}[1]{^{\@fnsymbol{#1}}}
\usepackage[accsupp]{axessibility}

\newcommand{\miou}{mIoU}
\newcommand\synteq{::=}
\newcommand\xbase{x'}
\newcommand\sparam{\alpha}
\newcommand{\reals}{\mathbb{R}}

\newcommand{\minisection}[1]{\vspace{0.04in} \noindent {\bf #1}\ \ }

\def\wacvPaperID{80} 

\wacvalgorithmstrack   

\wacvfinalcopy 

\newif\ifpreprint
\preprinttrue

\ifwacvfinal
    \ifpreprint
    \usepackage[pagebackref=true,breaklinks=true,colorlinks,bookmarks=false]{hyperref}
    \else
    \usepackage[breaklinks=true,bookmarks=false]{hyperref}
    \fi
\else
\usepackage[pagebackref=true,breaklinks=true,colorlinks,bookmarks=false]{hyperref}
\fi

\usepackage[sort&compress,capitalize,noabbrev,nameinlink]{cleveref}
\creflabelformat{equation}{#2#1#3}

\newcommand{\voc}[1]{
\begin{table*}[!h]
\caption{#1}
\centering
\vskip0.2cm
\begin{tabular}{c|cc|c|cc|c|cc|c|cc|c}
  & \multicolumn{3}{c|}{15-5 (2 steps)} & \multicolumn{3}{c|}{15-1 (6 steps)} & \multicolumn{3}{c|}{5-3 (6 steps)} & \multicolumn{3}{c}{10-1 (11 steps)}  \\

Method & 0-15 & 16-20 & all & 0-15 & 16-20 & all & 0-5 & 6-20 & all & 0-10 & 11-20 & all \\
 \hline
ILT$^{\dagger}$~\cite{michieli2019incremental} & 67.8 & 40.6 & 61.3 & 9.6 & 7.8 & 9.2 & 22.5 & 31.7 & 29.0 & 7.2 & 3.7 & 5.5 \\
SDR$^{\dagger}$~\cite{michieli2021continual} & 76.3 & 50.2 & 70.1 & 47.3 & 14.7 & 39.5 & - & - & - & 32.4 & 17.1 & 25.1 \\
PLOP~\cite{douillard2021plop} & 75.7 & 51.7 & 70.1 & 65.1 & 21.1 & 54.6 & 17.5 & 19.2 & 18.7 & 44.0 & 15.5 & 30.5 \\
MiB+UCD~\cite{yang2022uncertainty} & 78.5 & 50.7 & 71.5 & 51.9 & 13.1 & 42.2 & - & - & - & 33.7 & 26.5 & 31.1 \\
RCIL$^{\dagger}$~\cite{zhang2022representation} & \textbf{78.8} & 52.0 & \textbf{72.4} & 70.6 & 23.7 & 59.4 & 59.3 & 33.8 & 41.1 & 55.4 & 15.1 & 34.3 \\

\hline
MiB~\cite{cermelli2020modeling} & 75.5 & 49.4 & 69.0 & 35.1 & 13.5 & 29.7 & - & - & - & 12.3 & 13.1 & 12.7 \\
MiB*~\cite{cermelli2020modeling} & 76.4 & 49.4 & 70.0 & 48.1 & 15.8 & 40.4 & 58.2 & 41.3 & 46.1 & 14.1 & 13.8 & 13.9 \\
MiB+AWT (Ours) & \underline{77.3} & \underline{\textbf{52.9}} & \underline{71.5} & \underline{59.1} & \underline{17.2} & \underline{49.1} & \underline{61.8} & \underline{45.9} & \underline{50.4} & \underline{33.2} & \underline{18.0} & \underline{26.0} \\

\hline
SSUL~\cite{cha2021ssul} & 77.8 & 50.1 & 71.2 & \textbf{77.3} & 36.6 & \textbf{67.6} & \textbf{72.4} & 50.7 & 56.9 & 71.3 & 46.0 & 59.3 \\
SSUL+AWT (Ours) & \underline{78.0} & \underline{50.2} & \underline{71.4} & 77.0 & \underline{\textbf{37.6}} & \textbf{67.6} & 71.6 & \underline{\textbf{51.4}} & \underline{\textbf{57.1}} & \underline{\textbf{73.1}} & \underline{\textbf{47.0}} & \underline{\textbf{60.7}} \\
\hline
Joint & 79.8 & 72.4 & 77.4 & 79.8 & 72.4 & 77.4 & 76.9 & 77.6 & 77.4 & 78.4 & 76.4 & 77.4 \\
\hline
\end{tabular}
\label{tab:voc}
\end{table*}
}

\newcommand{\ade}[1]{
\begin{table*}[!h]
\caption{#1}
\centering
\vskip0.2cm
\resizebox{\textwidth}{!}{
\begin{tabular}{c|cc|c|cccccc|c|ccc|c}
  & \multicolumn{3}{c|}{100-50 (2 steps)} & \multicolumn{7}{c|}{100-10 (6 steps)} & \multicolumn{4}{c}{50-50 (3 steps)} \\

Method & 1-100 & 101-150 & all & 1-100 & 101-110 & 111-120 & 121-130 & 131-140 & 141-150 & all & 1-50 & 51-100 & 101-150 & all \\
 \hline
ILT$^{\dagger}$ \cite{michieli2019incremental} & 18.3 & 14.8 & 17.0 & 0.1 & 0.0 & 0.1 & 0.9 & 4.1 & 9.3 & 1.1 & 13.6 & 12.3 & 0.0 & 9.7 \\ 
PLOP \cite{douillard2021plop} & 41.9 & 14.9 & 32.9 & 40.6 & 15.2 & 16.9 & 18.7 & 11.9 & 7.9 & 31.6 & \textbf{48.6} & 30.0 & 13.1 & 30.4 \\
PLOP+UCD \cite{yang2022uncertainty} & 42.1 & 15.8 & 33.3 & \textbf{40.8} & - & - & - & - & - & 32.3 & 47.1 & - & - & 31.8 \\
SSUL* \cite{cha2021ssul} & 38.0 & 20.5 & 32.2 & 36.5 & \textbf{16.5} & 29.0 & 21.7 & 16.4 & 13.5 & 30.8 & 44.1 & 23.0 & 18.6 & 28.7 \\
RCIL$^{\dagger}$ \cite{zhang2022representation} & \textbf{42.3} & 18.8 & 34.5 & 39.3 & 14.6 & 26.3 & 23.2 & 12.1 & 11.8 & 32.1 & 48.3 & \textbf{31.3} & 18.7 & 32.5 \\

\hline
MiB$^{\dagger}$ \cite{cermelli2020modeling} & 40.5 & 17.7 & 32.8 & 38.3 & 12.6 & 10.6 & 8.7 & 9.5 & 15.1 & 29.2 & 45.3 & 26.1 & 17.1 & 29.3 \\
MiB*~\cite{cermelli2020modeling} & 41.5 & 22.9 & 35.3 & 38.9 & 10.3 & 13.8 & 12.3 & 5.1 & 13.0 & 29.6 & 46.1 & 27.1 & 21.8 & 31.8 \\
MiB+AWT (Ours) & 40.9 & \underline{\textbf{24.7}} & \underline{\textbf{35.6}} & \underline{39.1} & \underline{14.3} & \underline{\textbf{31.9}} & \underline{\textbf{24.4}} & \underline{\textbf{20.6}} & \underline{\textbf{15.2}} & \underline{\textbf{33.2}} & \underline{46.6} & \underline{30.1} & \underline{\textbf{23.6}} & \underline{\textbf{33.5}} \\
\hline
Joint & 44.3 & 28.2 & 38.9 & 44.3 & 26.1 & 42.8 & 26.7 & 28.1 & 17.3 & 38.9 & 51.1 & 38.3 & 28.2 & 38.9 \\
\hline
\end{tabular}}
\label{tab:ade20k}
\end{table*}
}

\newcommand{\secondade}[1]{
\begin{table}[t]
\caption{#1}
\centering
\vskip0.2cm
\begin{tabular}{c|cc|c}
  & \multicolumn{3}{c}{100-5 (11 steps)} \\

Method & 1-100 & 101-150 & all \\
 \hline
ILT$^{\dagger}$~\cite{michieli2019incremental} & 0.1 & 1.3 & 0.5 \\ 
PLOP~\cite{douillard2021plop} & \textbf{39.1} & 7.8 & 28.8 \\
RCIL$^{\dagger}$~\cite{zhang2022representation} & 38.5 & 11.5 & 29.6 \\
SSUL*~\cite{cha2021ssul} & 36.0 & \textbf{18.2} & 30.1 \\

\hline
MiB$^{\dagger}$~\cite{cermelli2020modeling} & 36.0 & 5.6 & 25.9 \\
MiB*~\cite{cermelli2020modeling} & 36.9 & 5.4 & 26.5 \\
MiB+AWT (Ours) & \underline{38.6} & \underline{16.0} & \textbf{31.1} \\
\hline
Joint & 44.3 & 28.2 & 38.9 \\
\hline
\end{tabular}
\label{tab:ade20k2}
\end{table}
}

\newcommand{\cityscapes}[1]{
\begin{table}[t]
\caption{#1}
\vskip0.2cm
\resizebox{\columnwidth}{!}{\begin{tabular}{c|cc|c|cc|c}
  & \multicolumn{3}{c|}{14-1 (6 steps)} & \multicolumn{3}{c}{10-1 (10 steps)} \\

Method & 1-14 & 15-19 & all & 1-10 & 11-19 & all \\
 \hline
FT & 0.0 & 10.1 & 2.5 & 0.0 & 4.8 & 2.2 \\
PLOP \cite{douillard2021plop} & 55.7 & 12.3 & 44.8 & \textbf{52.2} & 24.1 & 39.6 \\
RCIL \cite{zhang2022representation} & 55.7 & 7.1 & 43.6 & 51.0 & 17.4 & 35.9 \\
\hline
SSUL \cite{cha2021ssul} & 43.2 & 33.0 & 40.7 & 38.6 & 38.1 & 38.3 \\
SSUL+AWT & \underline{43.9} & \underline{\textbf{35.1}} & \underline{41.5} & 38.6 & \underline{\textbf{39.0}} & \underline{38.8} \\
\hline
MiB \cite{cermelli2020modeling} & \textbf{56.3} & 12.5 & 45.4 & 51.6 & 30.1 & 41.9 \\
MiB+AWT & 55.9 & \underline{19.8} & \underline{\textbf{46.9}} & 51.2 & \underline{37.2} & \underline{\textbf{44.9}} \\
\hline

Joint & 56.7 & 54.3 & 56.1 & 51.7 & 61.4 & 56.1 \\
\hline
\end{tabular}}
\label{tab:cityscapes}
\end{table}
}

\newcommand{\addablate}[1]{
\begin{table}[t]
\caption{#1}
\resizebox{\columnwidth}{!}{
\begin{tabular}{c|c|c|c|ccc}
 \multicolumn{4}{c|}{} & \multicolumn{3}{c}{VOC (15-1)} \\ 
Strategy & \% of filters & copy & add & 0-15 & 16-20 & all \\ 
 \hline
No transfer & 0 & $\times$ & $\times$ & 45.7 & 5.3 & 36.1 \\ 
MiB \cite{cermelli2020modeling} & 100 & $\checkmark$ & $\times$ & 48.1 & 15.8 & 40.4 \\
Random & 25 & $\times$ & $\checkmark$ & 46.3 & 6.1 & 36.8 \\ 
AWT & 25 & $\checkmark$ & $\times$ & 58.3 & 14.8 & 47.9 \\
AWT & 25 & $\times$ & $\checkmark$ & \textbf{59.1} & \textbf{17.2} & \textbf{49.1} \\
\end{tabular}}
\label{tab:addablate}
\end{table}
}

\newcommand{\ablate}[1]{
\begin{table}[t]
\centering
\caption{#1}
\begin{tabular}{c|ccc}
 & \multicolumn{3}{c}{VOC (15-1)}  \\
MiB+AWT & 0-15 & 16-20 & all  \\ 
 \hline
Max-Pool $\Longrightarrow$ Mean & 55.2 & 14.5 & 45.5 \\
Avg-Pool $\Longrightarrow$ Mean & 58.3 & 14.1 & 47.8 \\
Mean $\Longrightarrow$ Avg-Pool & 57.6 & 14.2 & 47.2 \\
Mean $\Longrightarrow$ Max-Pool & \textbf{59.1} & \textbf{17.2} & \textbf{49.1} \\
\end{tabular}
\label{tab:ablate}
\end{table}
}

\newcommand{\adecombo}[1]{
\begin{figure}[t]
     \centering
     \begin{subfigure}[b]{0.24\columnwidth}
         \centering
         \begin{subfigure}[b]{\columnwidth}
             \centering
             \includegraphics[width=\columnwidth]{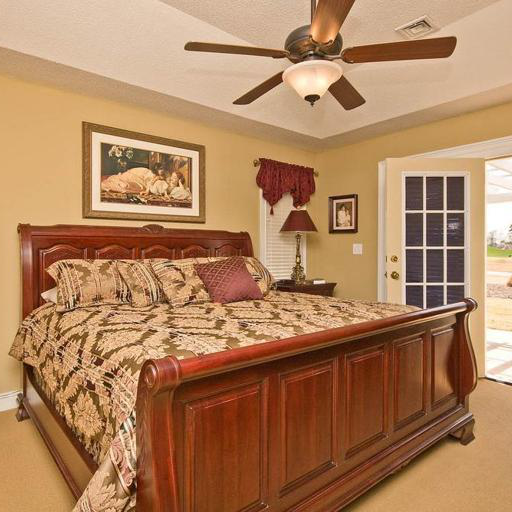}
             \label{fig:image1}
             \vspace{-1em}
         \end{subfigure}
         
         \begin{subfigure}[b]{\columnwidth}
             \centering
             \includegraphics[width=\columnwidth]{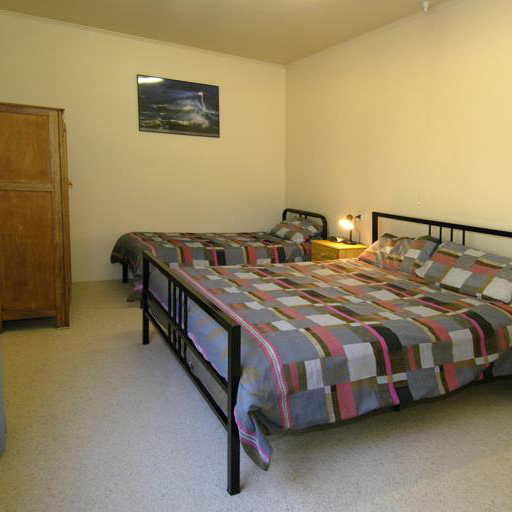}
             \label{fig:image2}
             \vspace{-1em}
         \end{subfigure}
         
         \begin{subfigure}[b]{\columnwidth}
             \centering
             \includegraphics[width=\columnwidth]{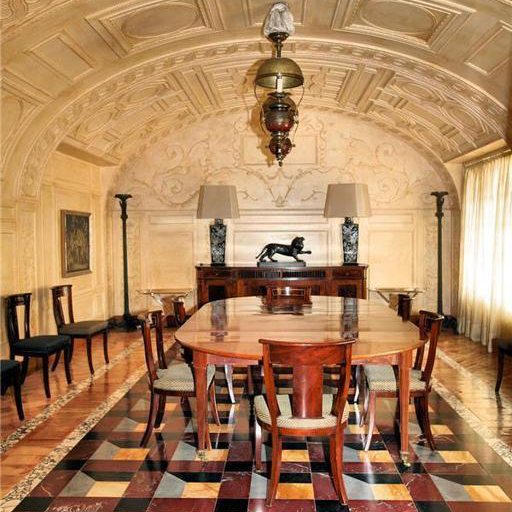}
             \label{fig:image3}
             \vspace{-1em}
         \end{subfigure}
         \caption{Image}
    \end{subfigure}
    \hfill
    \begin{subfigure}[b]{0.24\columnwidth}
         \centering
         \vspace{-1em}
         \begin{subfigure}[b]{\columnwidth}
             \centering
             \includegraphics[width=\columnwidth]{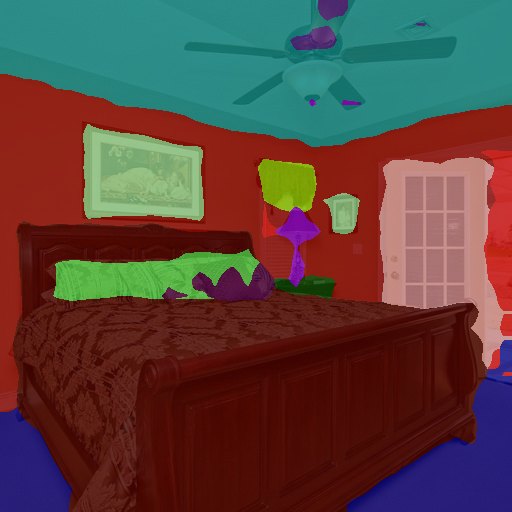}
             \label{fig:mib1}
             \vspace{-1em}
         \end{subfigure}

         \begin{subfigure}[b]{\columnwidth}
             \centering
             \includegraphics[width=\columnwidth]{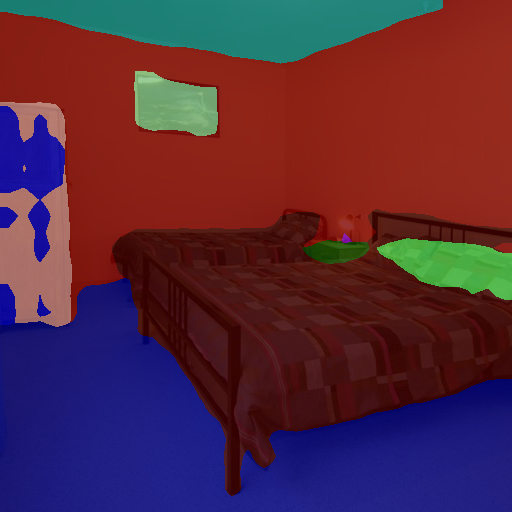}
             \label{fig:mib2}
             \vspace{-1em}
         \end{subfigure}

         \begin{subfigure}[b]{\columnwidth}
             \centering
             \includegraphics[width=\columnwidth]{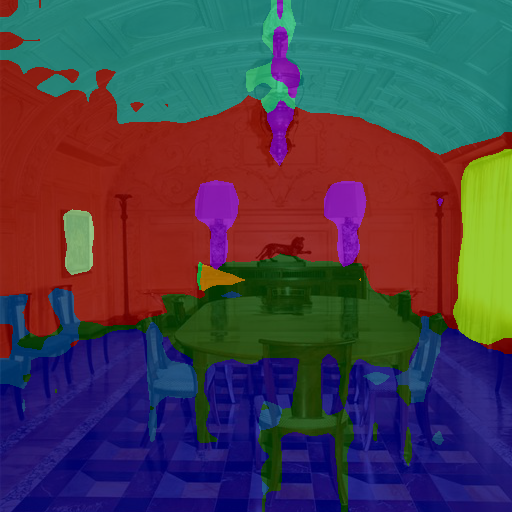}
             \label{fig:mib3}
             \vspace{-1em}
         \end{subfigure}

         \caption{MiB~\cite{cermelli2020modeling}}
    \end{subfigure}
    \hfill
    \begin{subfigure}[b]{0.24\columnwidth}
         \centering
         \vspace{-1em}
         \begin{subfigure}[b]{\columnwidth}
             \centering
             \includegraphics[width=\columnwidth]{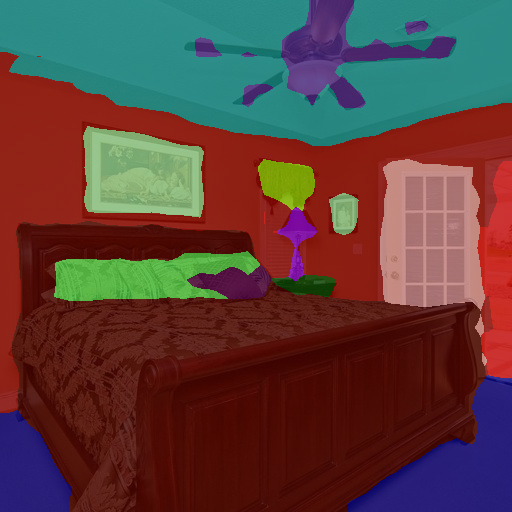}
             \label{fig:mib_awt1}
             \vspace{-1em}
         \end{subfigure}

         \begin{subfigure}[b]{\columnwidth}
             \centering
             \includegraphics[width=\columnwidth]{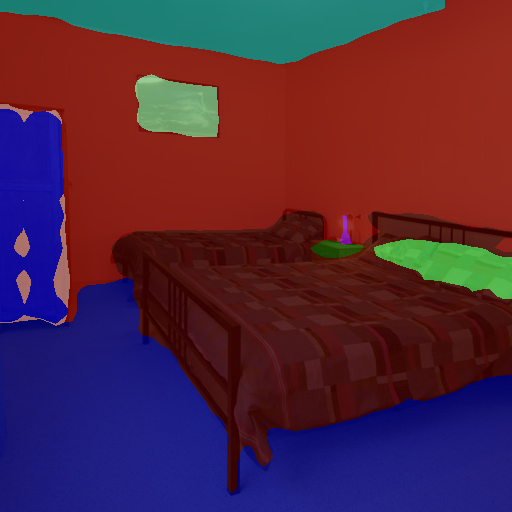}
             \label{fig:mib_awt2}
             \vspace{-1em}
         \end{subfigure}

         \begin{subfigure}[b]{\columnwidth}
             \centering
             \includegraphics[width=\columnwidth]{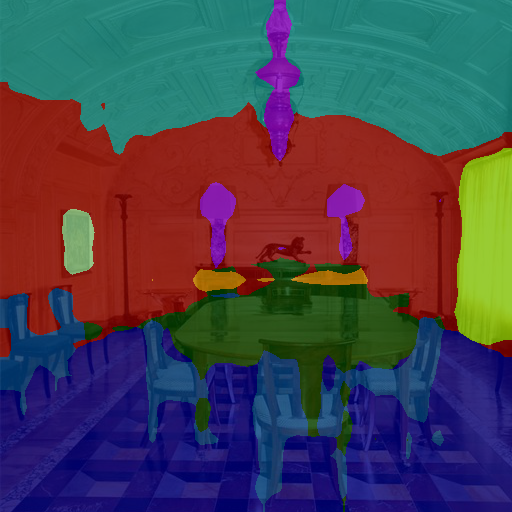}
             \label{fig:mib_awt3}
             \vspace{-1em}
     \end{subfigure}
     \caption{MiB+AWT}
    \end{subfigure}
    \hfill
    \begin{subfigure}[b]{0.24\columnwidth}
         \centering
         \vspace{-1em}
         \begin{subfigure}[b]{\columnwidth}
             \centering
             \includegraphics[width=\columnwidth]{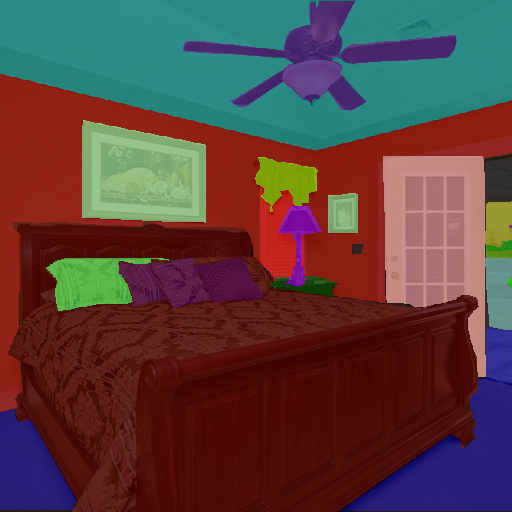}
             \label{fig:gt1}
             \vspace{-1em}
         \end{subfigure}

         \begin{subfigure}[b]{\columnwidth}
             \centering
             \includegraphics[width=\columnwidth]{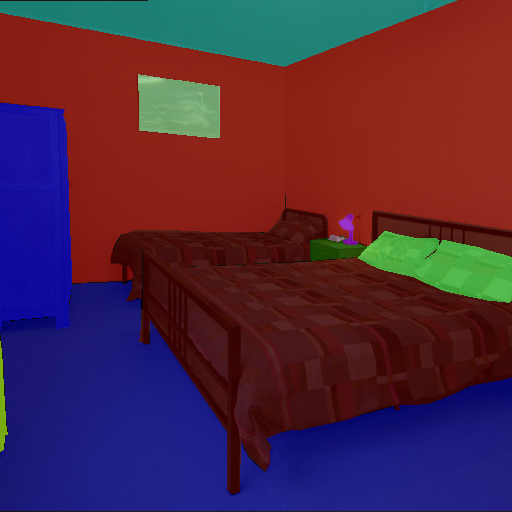}
             \label{fig:gt2}
             \vspace{-1em}
         \end{subfigure}

         \begin{subfigure}[b]{\columnwidth}
             \centering
             \includegraphics[width=\columnwidth]{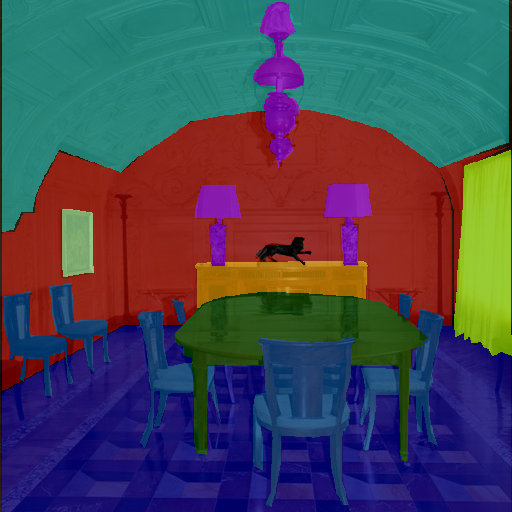}
             \label{fig:gt3}
             \vspace{-1em}
         \end{subfigure}

         \caption{GT}
    \end{subfigure}
    \caption{Visualization of predictions using MiB and MiB+AWT in 100-5 setting on test images of ADE20K.}
    \label{fig:ade_combo}
\end{figure}}

\newcommand{\initablate}[1]{
\begin{table}[t]
\caption{#1}
\resizebox{\columnwidth}{!}{
\begin{tabular}{cc|ccc}
 & & \multicolumn{3}{c}{VOC (15-1)}  \\
New Classifier Init & Iterations & 0-15 & 16-20 & all \\ 
 \hline
Random & $\times$ 1 & 45.7 & 5.3 & 36.1 \\ 
Random & $\times$ 2 & 39.7 & 6.6 & 31.8 \\ 
Random & $\times$ 4 & 29.9 & 7.5 & 24.6 \\ 
Weight transfer - MiB \cite{cermelli2020modeling} & $\times$ 1 & 48.1 & 15.8 & 40.4 \\
Weight transfer - AWT (Ours) & $\times$ 1 & \textbf{59.1} & \textbf{17.2} & \textbf{49.1} \\
\end{tabular}}
\label{tab:initablate}
\end{table}
}

\newcommand{\thresholdablate}[1]{
\begin{table}[t]
\centering
\caption{#1}
\begin{tabular}{c|ccc}
 & \multicolumn{3}{c}{VOC (15-1)}  \\
 Threshold for channel selection & 0-15 & 16-20 & all \\ 
 \hline
Top 10\% & 51.0 & 11.0 & 41.5 \\
Top 25\% & \textbf{59.1} & \textbf{17.2} & \textbf{49.1} \\
Top 50\% & 58.3 & 17.6 & 48.6 \\
Top 75\% & 56.8 & 14.9 & 46.8 \\
\end{tabular}
\label{tab:thresholdablate}
\end{table}
}

\newcommand{\voccombo}[1]{
\begin{figure*}[t]
     \centering
     \begin{subfigure}[b]{0.16\textwidth}
         \centering
         \begin{subfigure}[b]{\textwidth}
             \centering
             \includegraphics[width=\textwidth]{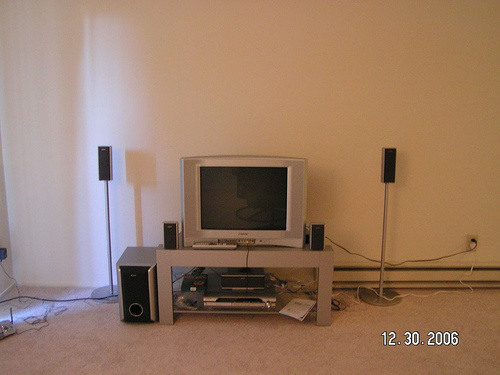}
             \vspace{-1em}
         \end{subfigure}
         
         \begin{subfigure}[b]{\textwidth}
             \centering
             \includegraphics[width=\textwidth]{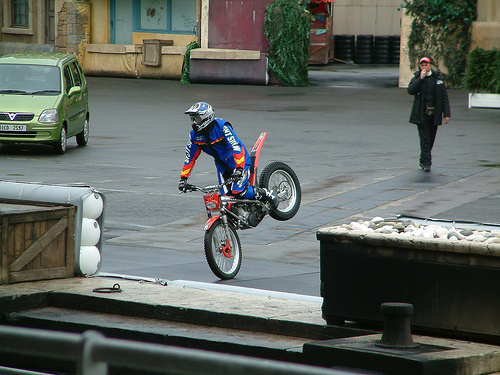}
             \vspace{-1em}
         \end{subfigure}
         
         \begin{subfigure}[b]{\textwidth}
             \centering
             \includegraphics[width=\textwidth]{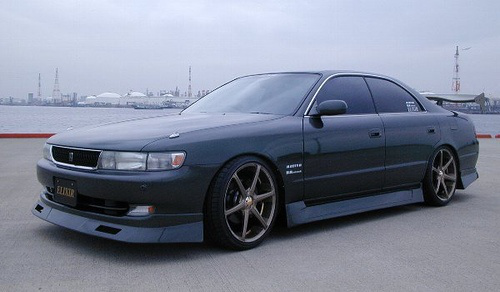}
             \vspace{-1em}
         \end{subfigure}
         
         \begin{subfigure}[b]{\textwidth}
             \centering
             \includegraphics[width=\textwidth]{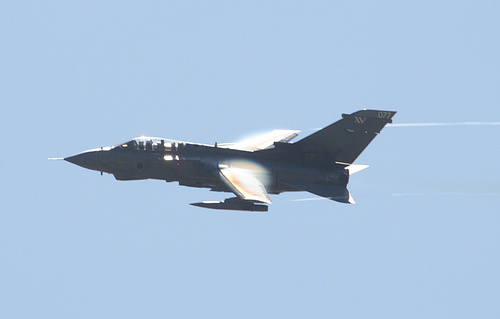}
             \vspace{-1em}
         \end{subfigure}
         
         \begin{subfigure}[b]{\textwidth}
             \centering
             \includegraphics[width=\textwidth]{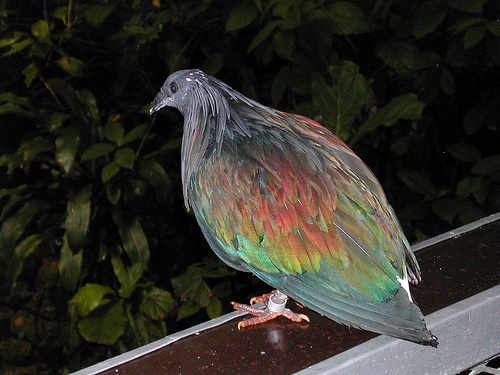}
             \vspace{-1em}
         \end{subfigure}
         
         \begin{subfigure}[b]{\textwidth}
             \centering
             \includegraphics[width=\textwidth]{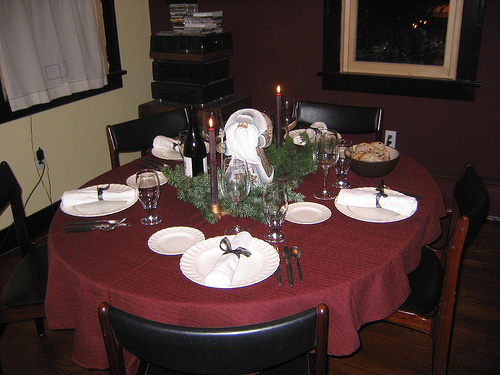}
             \vspace{-1em}
         \end{subfigure}
         
         \begin{subfigure}[b]{\textwidth}
             \centering
             \includegraphics[width=\textwidth]{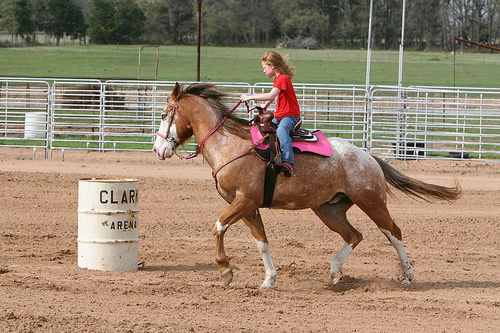}
             \vspace{-1em}
         \end{subfigure}
         
         \begin{subfigure}[b]{\textwidth}
             \centering
             \includegraphics[width=\textwidth]{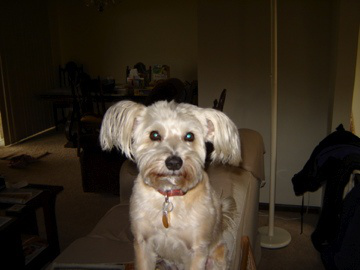}
             \vspace{-1em}
         \end{subfigure}
         
         \caption{Image}
    \end{subfigure}
    \hfill
    \begin{subfigure}[b]{0.16\textwidth}
         \centering
         \vspace{-1em}
         \begin{subfigure}[b]{\textwidth}
             \centering
             \includegraphics[width=\textwidth]{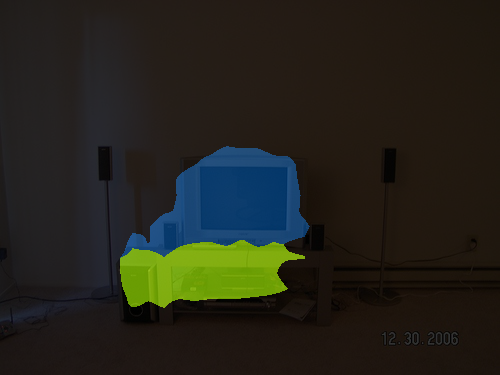}
             \vspace{-1em}
         \end{subfigure}

         \begin{subfigure}[b]{\textwidth}
             \centering
             \includegraphics[width=\textwidth]{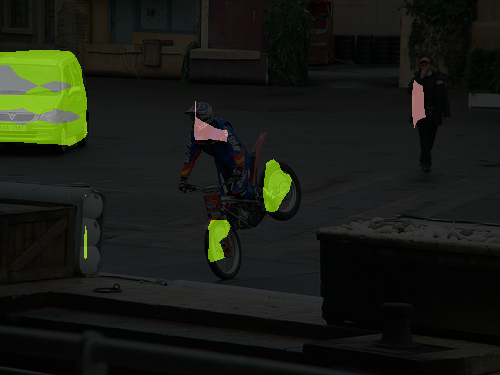}
             \vspace{-1em}
         \end{subfigure}

         \begin{subfigure}[b]{\textwidth}
             \centering
             \includegraphics[width=\textwidth]{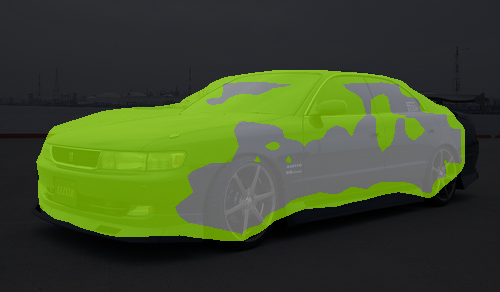}
             \vspace{-1em}
         \end{subfigure}
         
         \begin{subfigure}[b]{\textwidth}
             \centering
             \includegraphics[width=\textwidth]{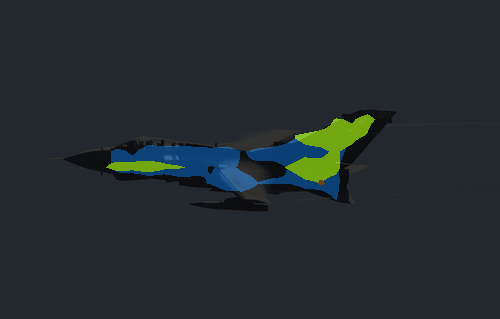}
             \vspace{-1em}
         \end{subfigure}
         
         \begin{subfigure}[b]{\textwidth}
             \centering
             \includegraphics[width=\textwidth]{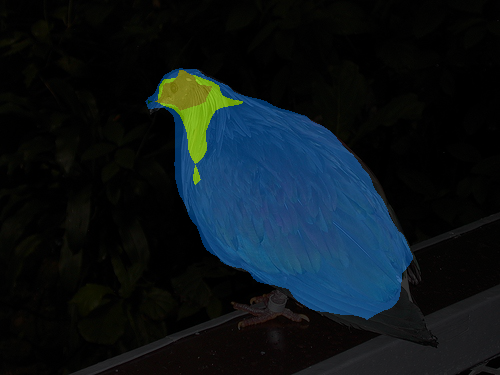}
             \vspace{-1em}
         \end{subfigure}
         
         \begin{subfigure}[b]{\textwidth}
             \centering
             \includegraphics[width=\textwidth]{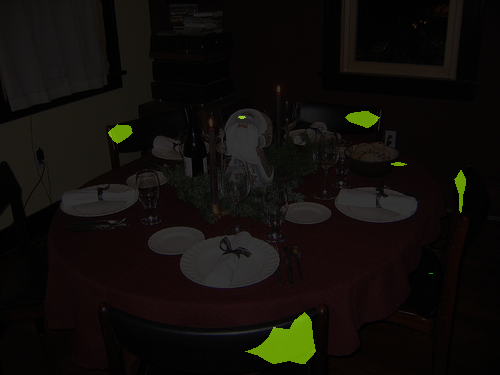}
             \vspace{-1em}
         \end{subfigure}
         
         \begin{subfigure}[b]{\textwidth}
             \centering
             \includegraphics[width=\textwidth]{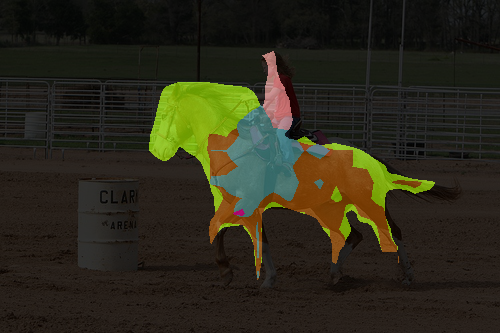}
             \vspace{-1em}
         \end{subfigure}
         
         \begin{subfigure}[b]{\textwidth}
             \centering
             \includegraphics[width=\textwidth]{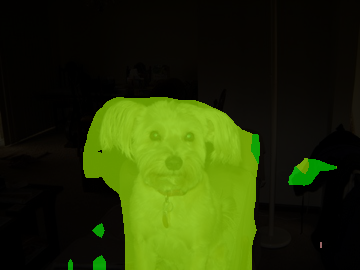}
             \vspace{-1em}
         \end{subfigure}

         \caption{MiB~\cite{cermelli2020modeling}}
    \end{subfigure}
    \hfill
    \begin{subfigure}[b]{0.16\textwidth}
         \centering
         \vspace{-1em}
         \begin{subfigure}[b]{\textwidth}
             \centering
             \includegraphics[width=\textwidth]{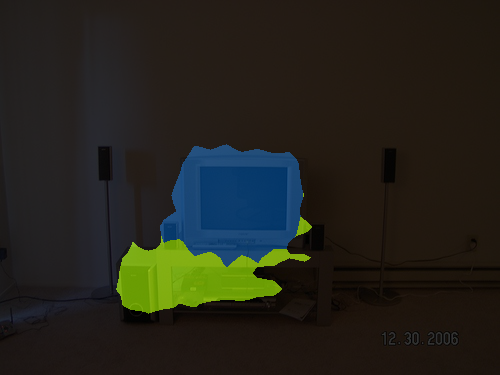}
             \vspace{-1em}
         \end{subfigure}

         \begin{subfigure}[b]{\textwidth}
             \centering
             \includegraphics[width=\textwidth]{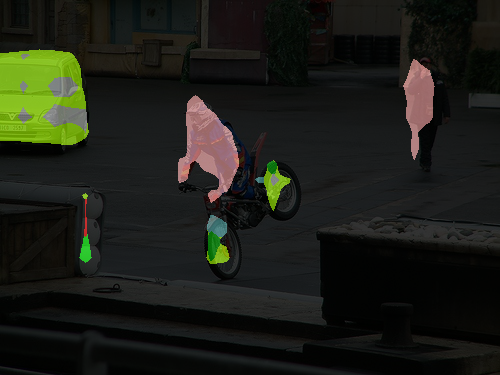}
             \vspace{-1em}
         \end{subfigure}

         \begin{subfigure}[b]{\textwidth}
             \centering
             \includegraphics[width=\textwidth]{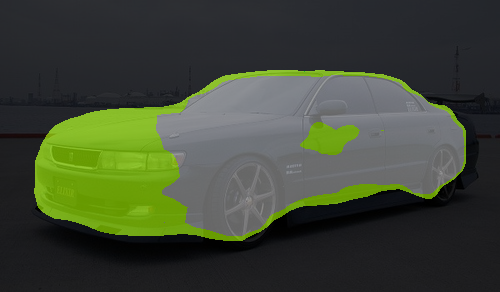}
             \vspace{-1em}
         \end{subfigure}

        \begin{subfigure}[b]{\textwidth}
             \centering
             \includegraphics[width=\textwidth]{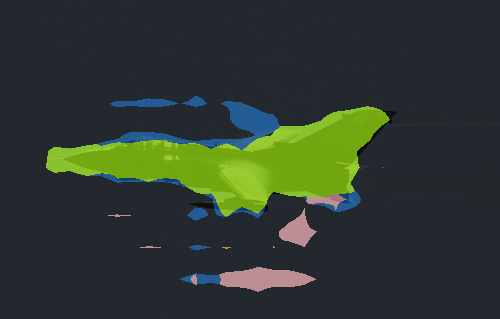}
             \vspace{-1em}
         \end{subfigure}
         
         \begin{subfigure}[b]{\textwidth}
             \centering
             \includegraphics[width=\textwidth]{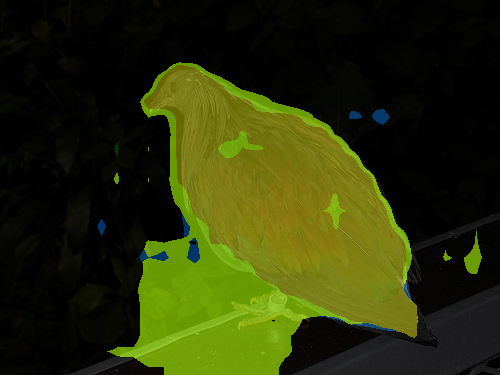}
             \vspace{-1em}
         \end{subfigure}
         
         \begin{subfigure}[b]{\textwidth}
             \centering
             \includegraphics[width=\textwidth]{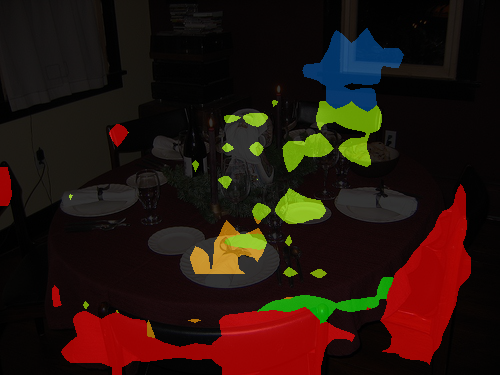}
             \vspace{-1em}
         \end{subfigure}
         
         \begin{subfigure}[b]{\textwidth}
             \centering
             \includegraphics[width=\textwidth]{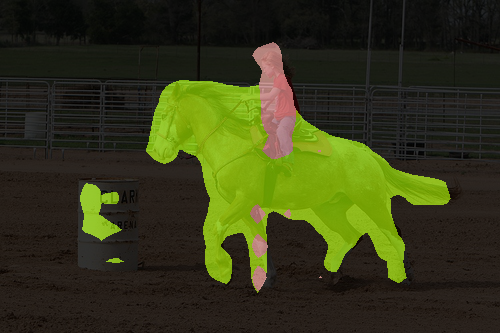}
             \vspace{-1em}
         \end{subfigure}
         
         \begin{subfigure}[b]{\textwidth}
             \centering
             \includegraphics[width=\textwidth]{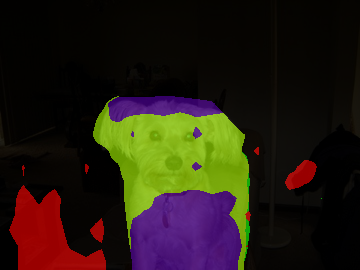}
             \vspace{-1em}
         \end{subfigure}
         
     \caption{MiB+AWT}
    \end{subfigure}
    \hfill
    \begin{subfigure}[b]{0.16\textwidth}
         \centering
         \vspace{-1em}
         \begin{subfigure}[b]{\textwidth}
             \centering
             \includegraphics[width=\textwidth]{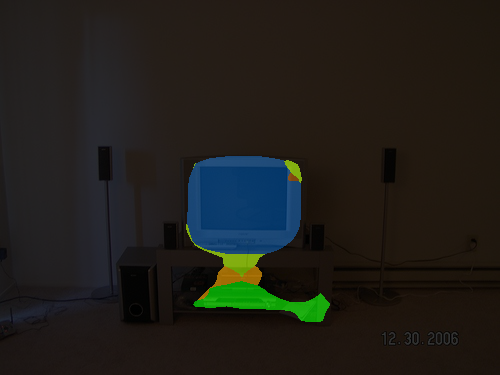}
             \vspace{-1em}
         \end{subfigure}

         \begin{subfigure}[b]{\textwidth}
             \centering
             \includegraphics[width=\textwidth]{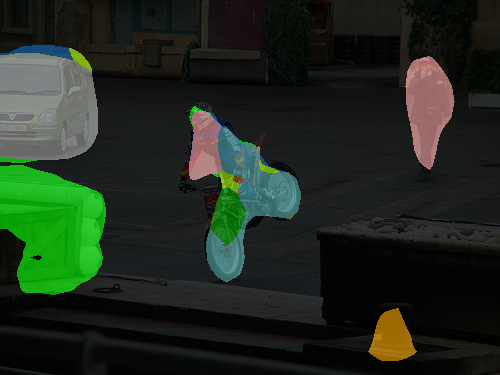}
             \vspace{-1em}
         \end{subfigure}

         \begin{subfigure}[b]{\textwidth}
             \centering
             \includegraphics[width=\textwidth]{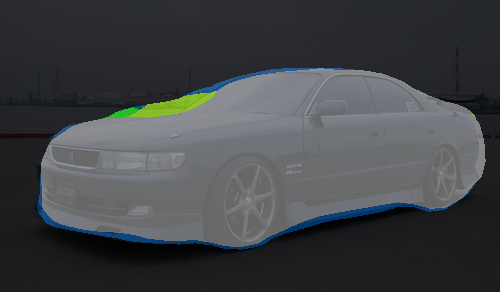}
             \vspace{-1em}
         \end{subfigure}
         
         \begin{subfigure}[b]{\textwidth}
             \centering
             \includegraphics[width=\textwidth]{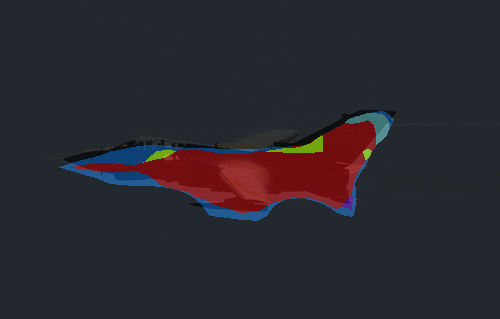}
             \vspace{-1em}
         \end{subfigure}
         
         \begin{subfigure}[b]{\textwidth}
             \centering
             \includegraphics[width=\textwidth]{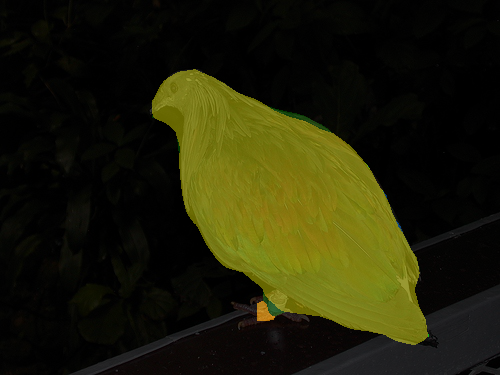}
             \vspace{-1em}
         \end{subfigure}
         
         \begin{subfigure}[b]{\textwidth}
             \centering
             \includegraphics[width=\textwidth]{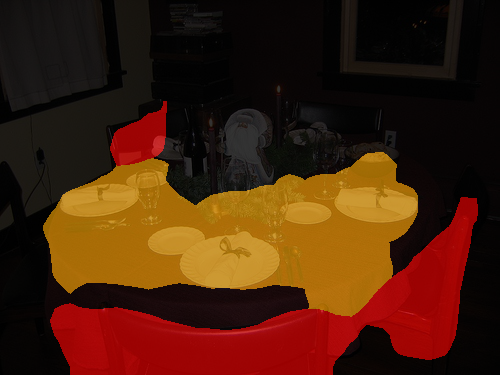}
             \vspace{-1em}
         \end{subfigure}
         
         \begin{subfigure}[b]{\textwidth}
             \centering
             \includegraphics[width=\textwidth]{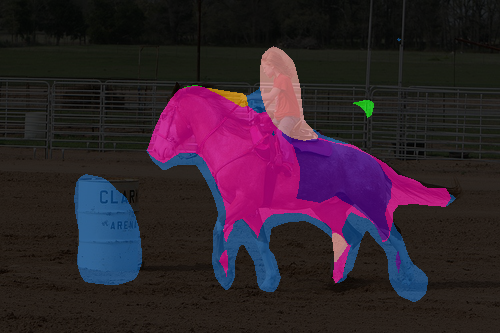}
             \vspace{-1em}
         \end{subfigure}
         
         \begin{subfigure}[b]{\textwidth}
             \centering
             \includegraphics[width=\textwidth]{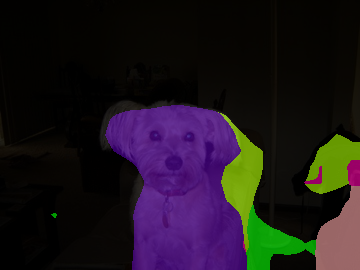}
             \vspace{-1em}
         \end{subfigure}

         \caption{SSUL~\cite{cha2021ssul}}
    \end{subfigure}
    \hfill
    \begin{subfigure}[b]{0.16\textwidth}
         \centering
         \vspace{-1em}
         \begin{subfigure}[b]{\textwidth}
             \centering
             \includegraphics[width=\textwidth]{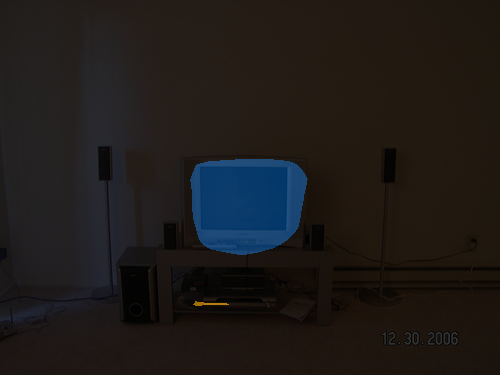}
             \vspace{-1em}
         \end{subfigure}

         \begin{subfigure}[b]{\textwidth}
             \centering
             \includegraphics[width=\textwidth]{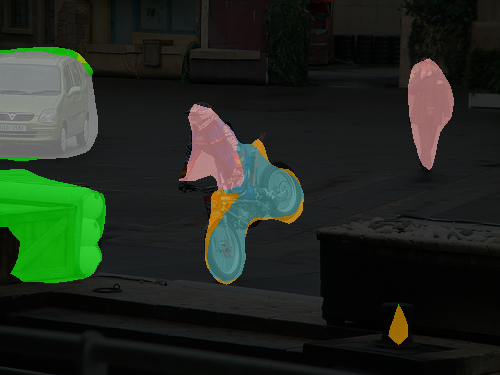}
             \vspace{-1em}
         \end{subfigure}

         \begin{subfigure}[b]{\textwidth}
             \centering
             \includegraphics[width=\textwidth]{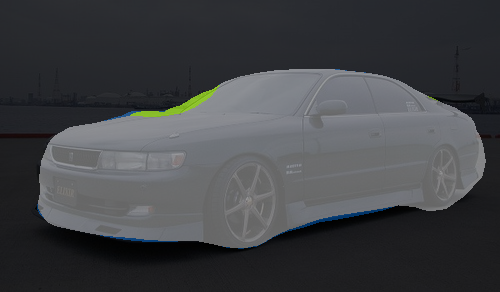}
             \vspace{-1em}
         \end{subfigure}
         
         \begin{subfigure}[b]{\textwidth}
             \centering
             \includegraphics[width=\textwidth]{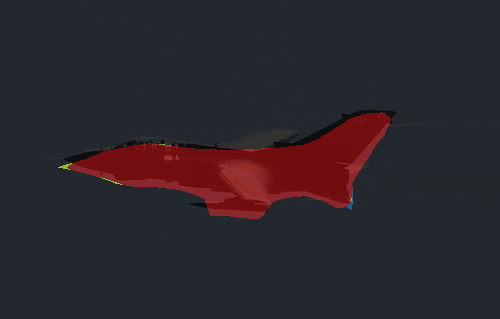}
             \vspace{-1em}
         \end{subfigure}
         
         \begin{subfigure}[b]{\textwidth}
             \centering
             \includegraphics[width=\textwidth]{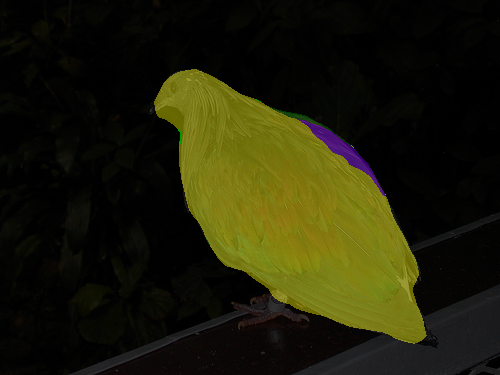}
             \vspace{-1em}
         \end{subfigure}
         
         \begin{subfigure}[b]{\textwidth}
             \centering
             \includegraphics[width=\textwidth]{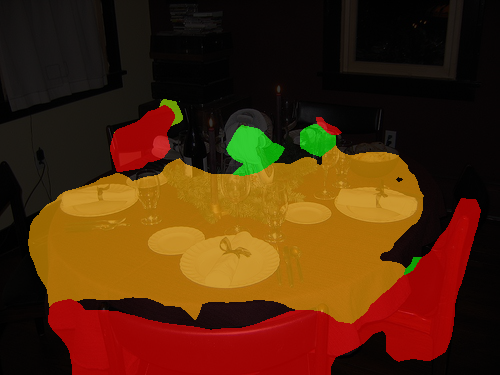}
             \vspace{-1em}
         \end{subfigure}
         
         \begin{subfigure}[b]{\textwidth}
             \centering
             \includegraphics[width=\textwidth]{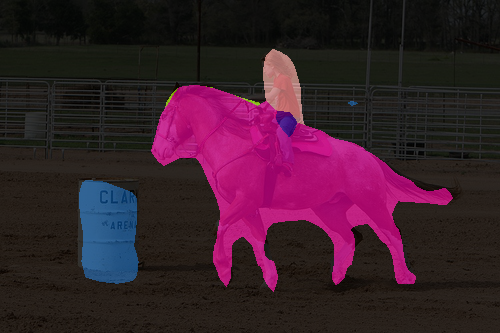}
             \vspace{-1em}
         \end{subfigure}
         
         \begin{subfigure}[b]{\textwidth}
             \centering
             \includegraphics[width=\textwidth]{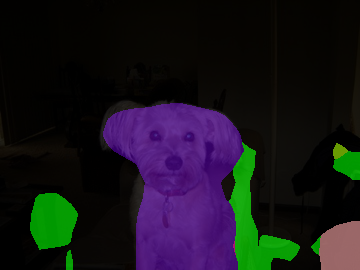}
             \vspace{-1em}
         \end{subfigure}

         \caption{SSUL+AWT}
    \end{subfigure}
    \hfill
    \begin{subfigure}[b]{0.16\textwidth}
         \centering
         \vspace{-1em}
         \begin{subfigure}[b]{\textwidth}
             \centering
             \includegraphics[width=\textwidth]{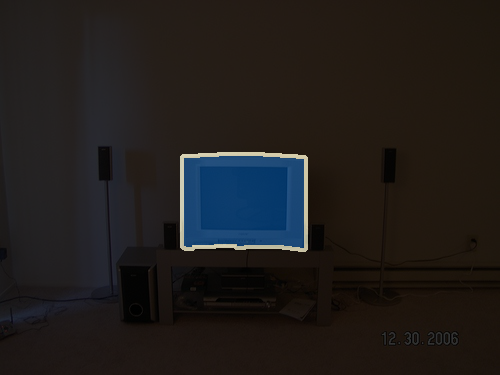}
             \vspace{-1em}
         \end{subfigure}

         \begin{subfigure}[b]{\textwidth}
             \centering
             \includegraphics[width=\textwidth]{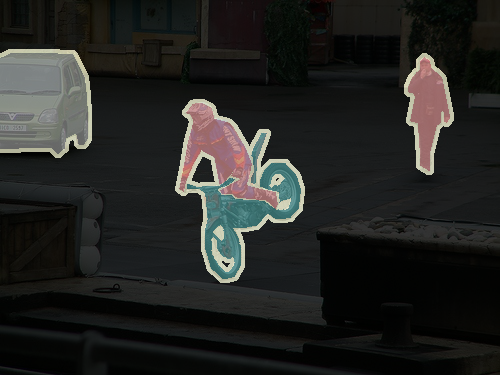}
             \vspace{-1em}
         \end{subfigure}

         \begin{subfigure}[b]{\textwidth}
             \centering
             \includegraphics[width=\textwidth]{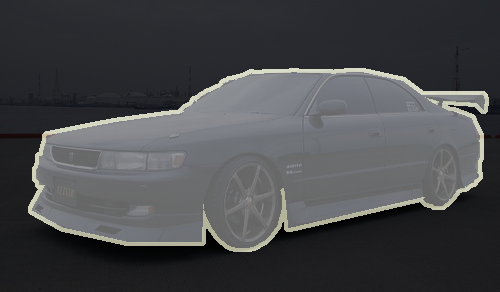}
             \vspace{-1em}
         \end{subfigure}
         
         \begin{subfigure}[b]{\textwidth}
             \centering
             \includegraphics[width=\textwidth]{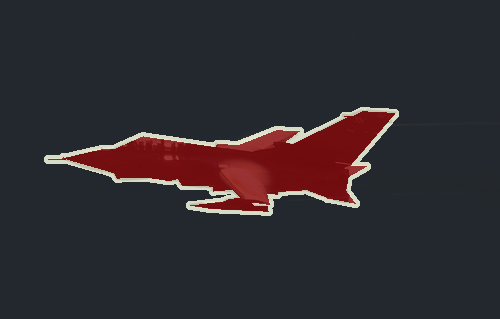}
             \vspace{-1em}
         \end{subfigure}
         
         \begin{subfigure}[b]{\textwidth}
             \centering
             \includegraphics[width=\textwidth]{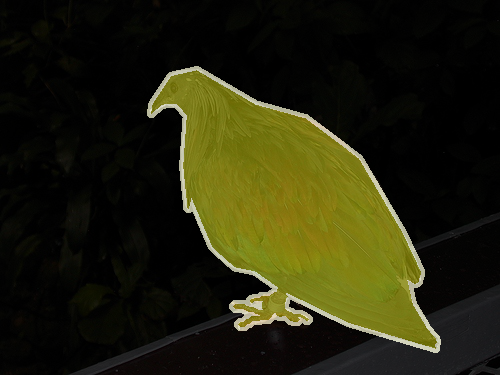}
             \vspace{-1em}
         \end{subfigure}
         
         \begin{subfigure}[b]{\textwidth}
             \centering
             \includegraphics[width=\textwidth]{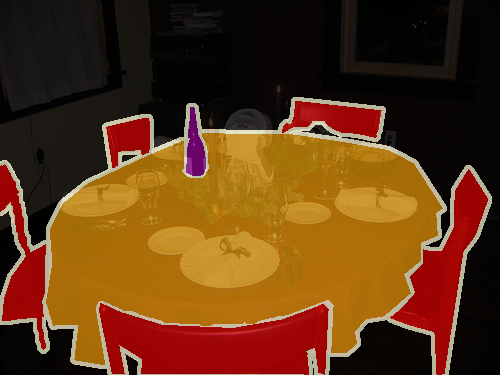}
             \vspace{-1em}
         \end{subfigure}
         
         \begin{subfigure}[b]{\textwidth}
             \centering
             \includegraphics[width=\textwidth]{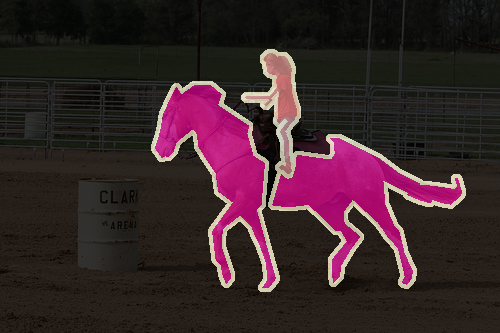}
             \vspace{-1em}
         \end{subfigure}
         
         \begin{subfigure}[b]{\textwidth}
             \centering
             \includegraphics[width=\textwidth]{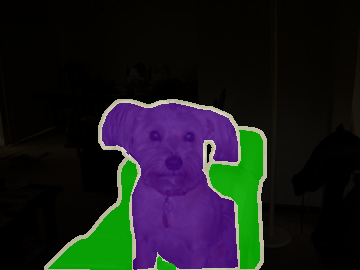}
             \vspace{-1em}
         \end{subfigure}

         \caption{GT}
    \end{subfigure}
    \caption{Visualization of predictions using MiB, MiB+AWT, SSUL and SSUL+AWT in 10-1 setting on test images of Pascal-VOC 2012.}
    \label{fig:voc_combo}
\end{figure*}}

\begin{document}

\title{Attribution-aware Weight Transfer: A Warm-Start\\ Initialization for Class-Incremental Semantic Segmentation}

\author{\quad Dipam Goswami$^{\dagger\mathsection}$ \enspace \qquad René Schuster$^{\dagger}$ \qquad Joost van de Weijer$^{\ddagger}$ \qquad Didier Stricker$^{\dagger}$ \\
{\tt\small dipamgoswami01@gmail.com} \enspace {\tt\small rene.schuster@dfki.de} \enspace {\tt\small joost@cvc.uab.es} \enspace {\tt\small didier.stricker@dfki.de} \\
$^{\dagger}$ DFKI - German Research Center for Artificial Intelligence, Kaiserslautern \\
         $^{\mathsection}$ Birla Institute of Technology and Science, Pilani \enspace \enspace
         $^{\ddagger}$ Computer Vision Center, Barcelona}

\maketitle
\thispagestyle{empty}

\begin{abstract}
In class-incremental semantic segmentation (CISS), deep learning architectures suffer from the critical problems of catastrophic forgetting and semantic background shift.
Although recent works focused on these issues, existing classifier initialization methods do not address the background shift problem and assign the same initialization weights to both background and new foreground class classifiers.
We propose to address the background shift with a novel classifier initialization method which employs gradient-based attribution to identify the most relevant weights for new classes from the classifier's weights for the previous background and transfers these weights to the new classifier.
This warm-start weight initialization provides a general solution applicable to several CISS methods. Furthermore, it accelerates learning of new classes while mitigating forgetting.
Our experiments demonstrate significant improvement in mIoU compared to the state-of-the-art CISS methods on the Pascal-VOC 2012, ADE20K and Cityscapes datasets.
\end{abstract}

\section{Introduction}
Semantic segmentation assigns a class label to every pixel of an image. The emergence of deep neural networks as well as the availability of pixel-level annotated datasets~\cite{everingham2015pascal, zhou2017scene, cordts2016cityscapes} has achieved state-of-the-art performance on semantic segmentation tasks~\cite{long2015fully, zhao2017pyramid, chen2018encoder}. The majority of papers in the field considers that all classes are labelled in all training data, and that all training data is jointly available. However, for many applications this is an unrealistic scenario, and the algorithm has to learn to segment all classes from partially labelled data, and every moment (called \emph{step} in CISS) only has access to a limited batch of training data. This restriction is imposed either by data storage limitations or data privacy and data security considerations~\cite{de2021continual}. Incremental learning~\cite{de2021continual, masana2020class} proposes algorithms for this setting where the main challenge is to prevent \emph{catastrophic forgetting}~\cite{mccloskey1989catastrophic} which refers to a drop in performance on classes learned in previous steps.

\begin{figure}[t]
\centering
\includegraphics[width=\columnwidth]{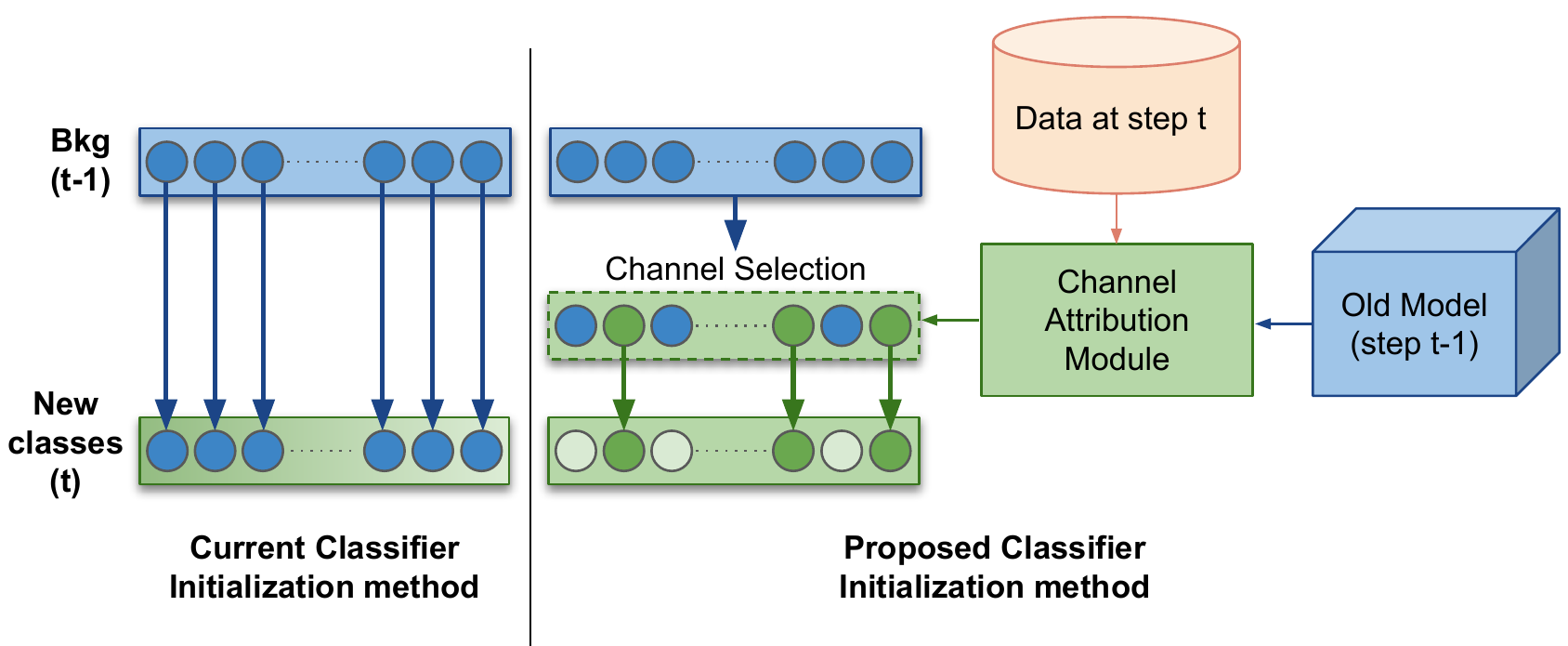}
\caption{Comparison of classifier initialization methods. Classifier initialization was previously found to be crucial to obtain good plasticity~\cite{cermelli2020modeling, douillard2021plop, cha2021ssul, yang2022uncertainty, zhang2022representation} of an incremental learner. However, the method did not address the \emph{background shift}. Previous methods (left) copy all filter weights from previous background (bkg) to initialize \emph{new classes}, our attribution-based weight selection (right) explicitly addresses the \emph{semantic background shift} by selecting only the relevant weights for the classification of \emph{new classes}. This allows us to combine the plasticity of the previous initialization method with a further reduction of catastrophic forgetting.}
\label{fig:compare}
\end{figure}

Another critical challenge faced by CISS approaches is the \emph{semantic background shift}. This challenge does not exist for incremental image classification, and is a result of the multi-class nature of image segmentation. 
The ground truth at any step provides labels for pixels belonging to current classes only and considers all other pixels as background while the model should correctly classify the old and current class pixels to their original labels and the future class pixels to the background.
So, the background class includes the real background class, the previously seen classes, and the future classes. 
As a result, there exists ambiguity due to changing definition of the background class from step to step.

In this paper, we highlight the importance of the initialization of the classifier's weights for the new classes. Since at every step, the final classifier layer has to predict the new classes in addition to the past classes, the classifier's weights for the new classes should be well aligned with the features extracted by the model. Thus, the initialization of classifier's weights is critical for stable training of the model and faster convergence on the new classes resulting in less forgetting of past classes. MiB~\cite{cermelli2020modeling} adapted weight imprinting~\cite{qi2018low} for segmentation and initialized the classifier's weights for the new classes and the background with the classifier's weights for the previous background. This initialization has been followed in most of the recent approaches~\cite{douillard2021plop, cha2021ssul, yang2022uncertainty, zhang2022representation} but it does not address the semantic background shift problem. Instead, we propose a new warm-start classifier initialization method that explicitly tackles the background shift, differentiating the current foreground classes from the previous background at the classifier level as depicted in \cref{fig:compare}.

We propose to transfer the learning of the future classes from the previous background to the new classes by weight transfer from relevant classifier input channels. 
The proposed method follows the strict incremental setting~\cite{masana2020class}, which does not need access to any data from past or future classes. Our method can be used with various CISS approaches. We validate our proposed approach using multiple CISS settings, class orders and ablation experiments. The main contributions can be summarized as follows:
\begin{itemize}
    \item To better address the background shift, we apply an attribution method to identify the most relevant classifier channels for predicting the new classes as previous background and transfer only those channel weights.
    \item Our method reduces catastrophic forgetting on old classes while improving plasticity on new classes, owing to quicker convergence on new classes.
    \item We experimentally show that our method significantly outperforms the state-of-the-art approaches on several incremental settings and datasets.
\end{itemize}

\section{Related Work} 

\noindent\textbf{Semantic Segmentation:}
Commonly used segmentation approaches are based on Fully Convolutional Networks (FCNs)~\cite{long2015fully}. These methods improve the accuracy by using multiscale representations~\cite{lin2017refinenet}, retaining more spatial information by using atrous convolution~\cite{chen2017deeplab} or convolution with upsampled filters, modelling contextual cues~\cite{chen2017deeplab}, or by using attention mechanisms~\cite{chen2016attention, zhao2018psanet}. Recent approaches used strip pooling~\cite{hou2020strip} along the width or height dimensions to capture both global and local statistics. In our work, we use the Deeplabv3~\cite{chen2017rethinking} architecture which employs atrous convolution in parallel manner in order to capture multi-scale context to segment objects at multiple scales.

\noindent\textbf{Incremental Learning:}
Most studies in incremental learning have focused on object detection and classification problems~\cite{castro2018end, li2017learning, rebuffi2017icarl, shmelkov2017incremental, wu2018incremental}. Some of these works use replay-based approaches, which store samples from previous tasks~\cite{rebuffi2017icarl, castro2018end} or generate training data~\cite{kemker2017fearnet, shin2017continual}. Some methods extend the initial architecture to learn new classes~\cite{yoon2018lifelong, li2019learn} or use embedding networks~\cite{Yu_2020_CVPR} or use classifier drift correction to account for changing class distributions~\cite{belouadah2019il2m, belouadah2020scail}. Distillation-based methods constrain the learning of the model on new tasks by penalizing updates on the weights~\cite{kirkpatrick2017overcoming, aljundi2018memory} or the gradients~\cite{lopez2017gradient, AGEM} or the intermediate features~\cite{hou2019learning, dhar2019learning, douillard2020podnet}. Our work focuses on the distillation-based approaches for semantic segmentation.

\noindent\textbf{Class-Incremental Semantic Segmentation:}
Recently, incremental learning has been studied for semantic segmentation~\cite{cermelli2020modeling, cha2021ssul, douillard2021plop, douillard2021tackling, kalb2021continual, michieli2021continual, klingner2020class, michieli2019incremental, yu2022self}. Initial approaches use relevant examples from old tasks and perform rehearsal for segmentation in medical imaging~\cite{ozdemir2019extending} and remote sensing data~\cite{tasar2019incremental}. Michieli \etal ~\cite{michieli2019incremental} considered an incremental setting where labels for old classes are available when learning new tasks. Cermelli \etal~\cite{cermelli2020modeling} was the first to highlight the semantic background shift and proposed a novel distillation method to tackle the shift. Douillard \etal~\cite{douillard2021plop, douillard2021tackling} proposed using multi-scale spatial distillation loss to preserve short and long range dependencies. Cha \etal~\cite{cha2021ssul} proposed SSUL which considers a separate class apart from the semantic background class for old and future classes in addition to freezing the backbone and past classifiers. UCD~\cite{yang2022uncertainty} proposed to enforce similarity between features for pixels of same classes and reduce feature similarity for pixels of different classes. RCIL~\cite{zhang2022representation} decoupled the learning of both old and new classes and introduced pooled cube knowledge distillation on channel and spatial dimensions. 

Replay of samples from previous classes has also been used for CISS either by storing images from old tasks~\cite{cha2021ssul} or by recreating them using generative networks~\cite{maracani2021recall}. Self-training approach using unlabelled data~\cite{yu2022self} has also been proposed.
We propose to model the semantic background shift for the classifier initialization used in~\cite{cermelli2020modeling, douillard2021plop, cha2021ssul, yang2022uncertainty, zhang2022representation} without using any data from the previous steps.

\noindent\textbf{Attribution Methods:}
Feature attribution methods assign importance scores to the features for a given input which are responsible for the class prediction. Existing attribution methods are based on perturbation or backpropagation. Perturbation methods~\cite{zeiler2014visualizing} compute the attributions of input features by removing or masking them and then do a forward pass to measure the difference in outputs. 
Backpropagation methods compute the attributions for the input features by doing one forward and backward pass. 
Some of these methods are DeepLIFT~\cite{shrikumar2017learning}, Integrated Gradients~\cite{sundararajan2017axiomatic} and Layer-wise Relevance Propagation (LRP)~\cite{bach2015pixel}. 
We use the popular Integrated Gradients~\cite{sundararajan2017axiomatic} which requires no modification to the network and is simple to implement.

\section{Proposed Weight Transfer Method}

\subsection{Class-Incremental Semantic Segmentation}

Consider an image $x$ and label space $\mathcal{C}$, semantic segmentation aims to assign a label $c_i \in \mathcal{C}$  
to every pixel in $x$. Provided with a training set $\mathcal{T}$, a model $f_{\theta}$ having parameters $\theta$ is learned which maps the input image to the pixel-wise class probabilities. In an incremental setup, the model is learned in $t = 1...T$ steps. The training set at incremental step $t$ is 
$\mathcal{T}^t = \{(x_1^t, y_1^t),...,(x_{n^t}^t, y_{n^t}^t)\}$ where $x_i^t \in X^t$ is the set of images, $y_i^t \in Y^t$
is the set of corresponding ground truth maps and a new set of classes $\mathcal{C}^t$ is added to the existing set of classes $\mathcal{C}^{1:t-1}$. Since the background class is present in all the incremental steps, we denote it as $b^t$ at step $t$. The model at step $t$ is denoted as $f_{\theta^t}$ which learns the parameters $\theta^t$. 

For an image $x_i^t \in X^t$, the ground truth segmentation map $y_i^t \in Y^t$ only provides the labels of current classes $\mathcal{C}^t$ while collapsing all other labels (old classes $\mathcal{C}^{1:t-1}$ and future classes $\mathcal{C}^{t+1:T}$) into the background class $b^t$. The model after step $t$ is expected to predict all classes learned over time $\mathcal{C}^{1:t}$. Here, both the real background class pixels and the future class $\mathcal{C}^{t+1:T}$ pixels should be predicted as background $b^t$. Hence, the future classes classified as background after the first step gradually become the foreground during the incremental steps. During the inference of the final step, only the real background class should be classified as the background. 

\subsection{Problems with Existing Initialization Method}
We discuss the existing initialization approach and the problems that are yet to be addressed. Since the pixels of $\mathcal{C}^t$ are learned as background $b^{t-1}$ at the previous step, the old model $f_{\theta^{t-1}}$ will most likely assign these pixels to class $b^{t-1}$. To account for this initial bias on predictions of $f_{\theta^t}$ for pixels of $\mathcal{C}^t$, Cermelli \etal,~\cite{cermelli2020modeling} proposed to initialize the classifier's weights for the classes in $\mathcal{C}^t$ (including background) with the weights for the previous background class so that the background class probability is uniformly spread among the classes in $\mathcal{C}^t$ ($b^t \in \mathcal{C}^t$). It is important to note, that this classifier initialization was found to be crucial to achieve good plasticity. For several settings, classifier initialization more than doubles performance on the classes learned after the first step (see for example Table 3 in~\cite{cermelli2020modeling}).

However, this direct transfer of classifier's weights from background to new classes does not directly address the shift of classes from background to foreground across time, which is one of the main challenge for CISS problems. The background classifier weights are learned for the real background and future classes $\mathcal{C}^{t+1:T}$ but the direct transfer guides the model to initially assign high probabilities for pixels of $\mathcal{C}^{t+1:T}$ and real background class to $\mathcal{C}^t$ instead of $b^t$.

\begin{figure}[t]
    \centering
    \includegraphics[width=0.95\columnwidth]{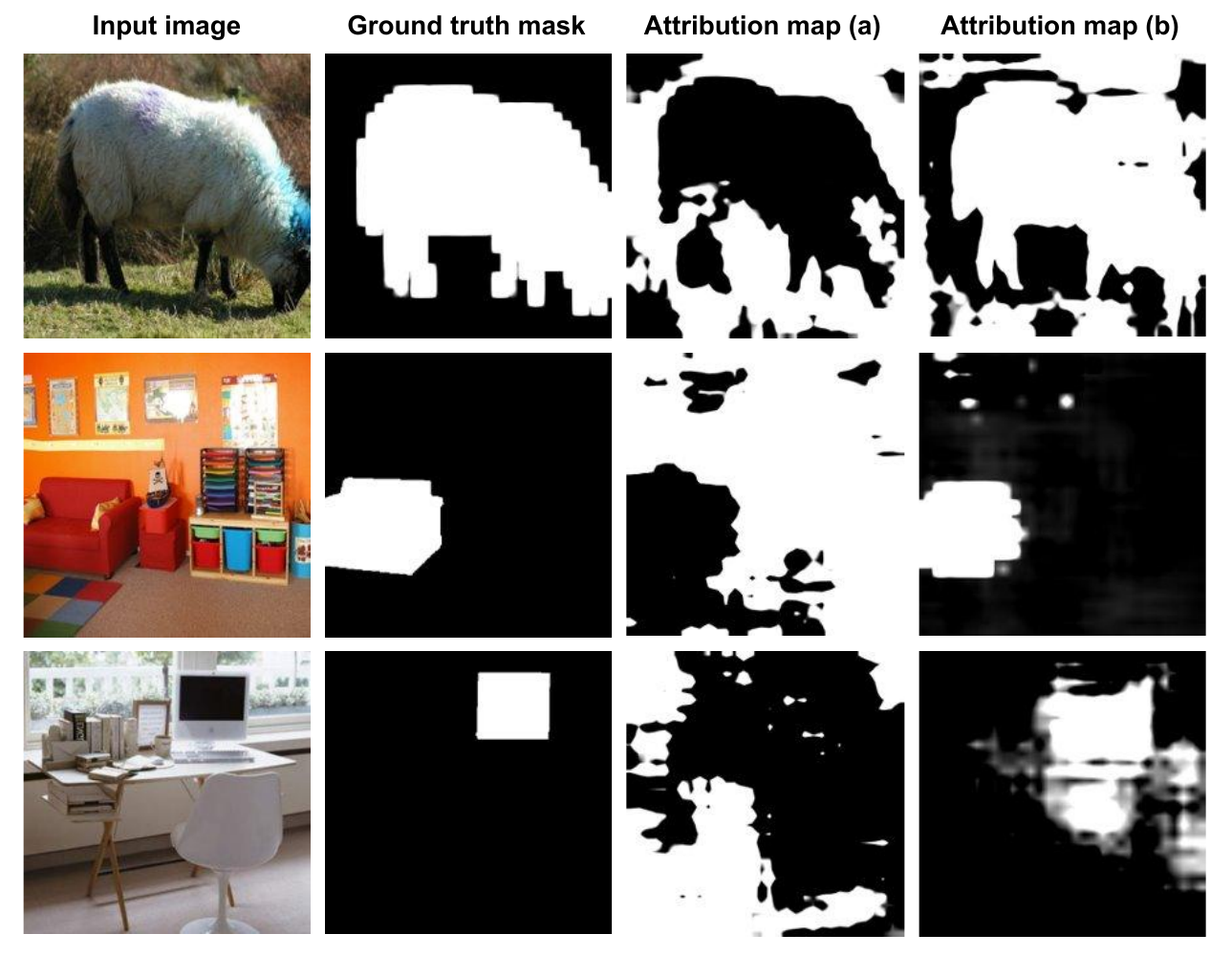}
    \caption{Attribution maps for the background (\emph{bkg}) class corresponding to different channels of the classifier layer. The new classes (sheep, sofa, monitor) belong to the \emph{bkg} of the previous step. Attribution map (a) has high contribution towards predicting the real \emph{bkg} pixels and does not predict the new class while the attribution map (b) contributes more towards predicting the new class pixels as \emph{bkg}.}
    \label{fig:attribution}
\end{figure}

\begin{figure*}[h]
    \centering
    \includegraphics[width=0.9\textwidth]{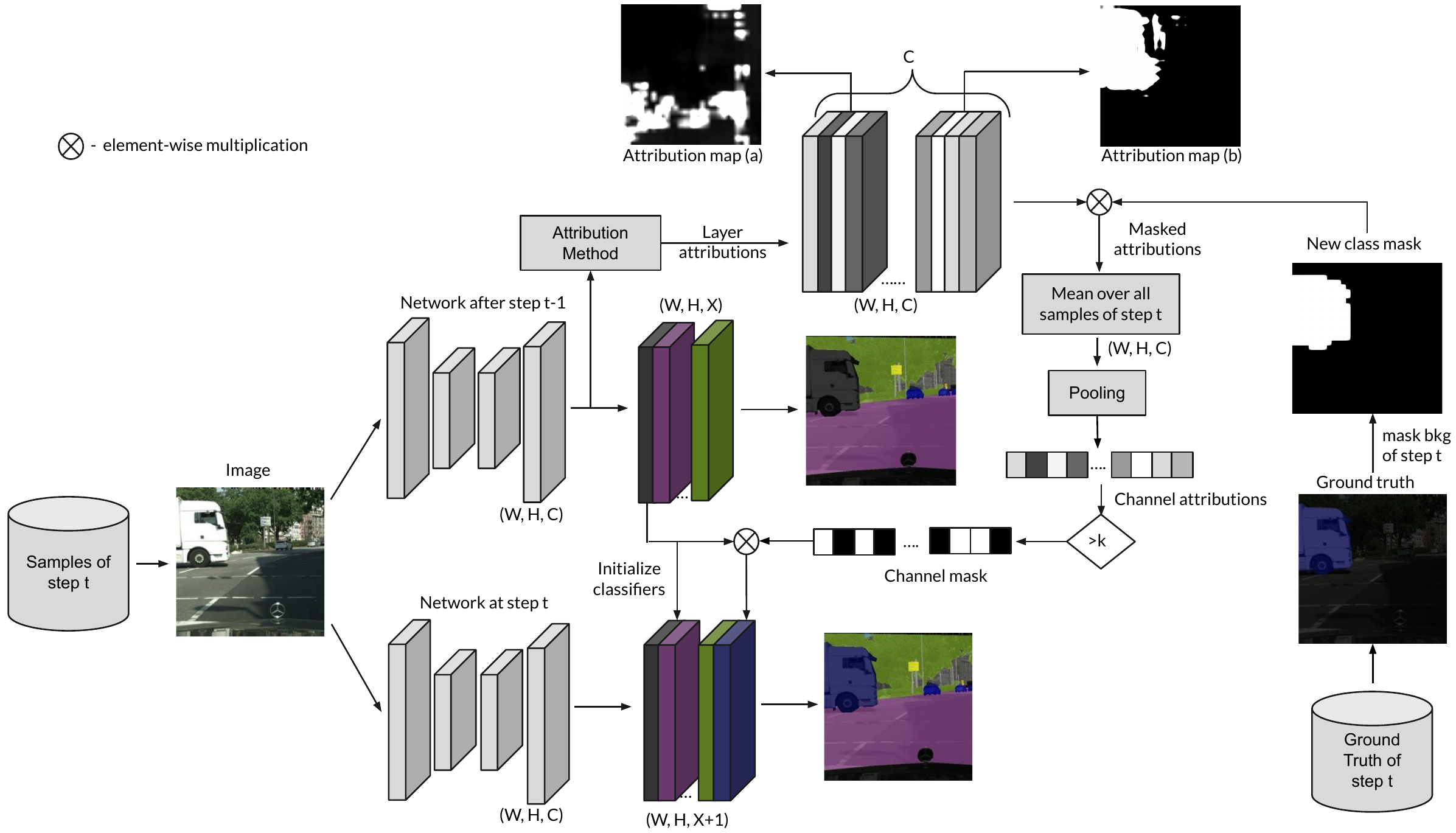}
    \caption[Novel Classifier Initialization method]{Overview of AWT: Images from the current step are given to the old model at step $t-1$. The inputs to the classifier layer are used to generate the layer attributions. Here, the attribution map (b) is more significant for new class pixels than map (a). We mask the attribution maps for the background pixels using the ground truth. Masked attributions from all images of step $t$ are averaged and max-pooled to obtain channel attributions. The significant channels are then selected using a threshold $k$ and these channel weights are transferred to the classifier weights for the new classes.}
    \label{fig:novelinit}
\end{figure*}

\subsection{Novel Warm-Start Classifier Initialization}

To explicitly address the background shift at the initialization stage, we propose Attribution-aware Weight Transfer (AWT) as illustrated in~\cref{fig:novelinit}. AWT aims to transfer only the significant classifier channel weights from the background $b^{t-1}$ to the new classes in $\mathcal{C}^t$. We use attribution methods to select the classifier weights for the background at the previous step, which have significant contributions in predicting the pixels of new classes in $\mathcal{C}^t$ as $b^{t-1}$ (as depicted in \cref{fig:attribution}). Here, we exploit the ability of the background classifier to learn different representations using different channels. This selection separates the classifier level weights for the new classes in $\mathcal{C}^t$ and the future classes. AWT aims not to transfer the significant classifier weights for the future classes $\mathcal{C}^{t+1:T}$ to the new classes thereby maintaining the stability of the model and accelerating learning of the new classes.

\subsubsection{Attribution-aware Channel Selection}
Integrated Gradients~\cite{sundararajan2017axiomatic} approximates the integral of the gradients of the model's output with respect to the inputs along a straight-line path from baselines to inputs. Here, baseline refers to the starting point from which integral is computed and is taken as blank (zero) input. We employ the Integrated Gradients attribution method, which assigns importance scores to the inputs of the classifier layer for predicting the background $b^{t-1}$. More details of Integrated Gradients are provided in the supplementary material.

We use the images from the current training set $X^t$ and the old model $f_{\theta^{t-1}}$ for computing the attribution maps for each of the input channels to the classifier layer. Since, the old and future class pixels are labelled as $b^t$ in the ground truth maps at step $t$, we obtain a ground truth mask $\bar{y}^t_{i}$ for new classes only by masking the pixels $m$ belonging to $b^t$ as follows: 
\begin{equation}
    {\bar y}^{t}_{i}[m] = \begin{cases}
      0 & \text{if $y^t_i[m] = b^t$} \\
      1 & \text{otherwise}
    \end{cases}
\end{equation}

We use the ground truth masks $\bar{y}^t_{i}$ to obtain the attributions corresponding to the pixels of new classes in $\mathcal{C}^t$ only. Since we consider only one set of channel mask for each new class in $\mathcal{C}^t$, we take the mean of the masked attributions from all the images in $X^t$. Let $A$ be the set of classifier layer attribution maps, each of size $W \times H \times C$ for all images in $X^t$ obtained using Integrated Gradients, then we obtain the mean attribution as follows:
\begin{equation}
    A_{avg}= \frac{1}{n^t} \sum_{i =1}^{n^t} A(x^t_i)\odot {\bar y}_{i}^t
\end{equation}
where $\odot$ refers to an element-wise multiplication along the spatial dimensions.

Max-Pooling is performed on the mean attribution $A_{avg}$ to get an attribution value for each of the channels of classifier's weights for background $b^{t-1}$. We transform the mean attribution map of size $W \times H \times C$ to channel attribution $A_c$ of size $C$ with pooling along the channel axis as follows:
\begin{equation}
    A_c[c] = \max_{w \in [1,W],h \in [1,H] }(A_{avg}[w, h, c])
\end{equation}
where $c \in [1,C]$. 
Note that we choose max-pooling over average-pooling based on experiments in \cref{sec:ablation}.

\subsubsection{Classifier Initialization}
A threshold $k$ is applied on the channel attribution to obtain a channel mask $c_{mask}$ to select the most contributing channels as follows:
\begin{equation}
    c_{mask}[c] = 
    \begin{cases}
      1 & \text{if $A_c[c] > k$} \\
      0 & \text{otherwise}
    \end{cases}
\end{equation}
where $A_c$ denotes the channel attribution and $c$ refers to the channels.
Let the classifier's weights for the class $c$ at step $t$ be $w^t_c$ and the classifier's weights for the background at step $t-1$ be $w^{t-1}_{b}$. We propose to initialize the classifier's weights for the new classes with the selected channel's weights as follows:
\begin{equation}
    \label{init}
    w^t_c = \begin{cases}
    w^{t*}_c + w^{t-1}_{b} \cdot c_{mask} & \text{if $c \in \mathcal{C}^t \setminus b^t$} \\
      w^{t-1}_c & \text{otherwise}
    \end{cases}       
\end{equation}
where $w^{t*}_c$ refers to the default initialized
weights. We transfer the masked weights by adding them on top of the default weights and thus we avoid having zero weights for the remaining channels. We show in \cref{sec:ablation} that adding the weights is beneficial compared to copying.

\voc{Experimental results on Pascal VOC 2012. Improvements using AWT \underline{underlined}. Best among columns in \textbf{bold}. $\dagger$:~results excerpted from~\cite{zhang2022representation}. * implies results come from re-implementation. Other results come from the respective papers.}

\section{Experiments}

\subsection{Experimental settings}

\noindent\textbf{Datasets:}
We conduct experiments on the segmentation datasets namely Pascal VOC 2012~\cite{everingham2015pascal}, ADE20K~\cite{zhou2017scene} and Cityscapes~\cite{cordts2016cityscapes} using different incremental splits. Pascal VOC 2012~\cite{everingham2015pascal} covers 20 object (or \emph{things}) classes and one background class. ADE20K~\cite{zhou2017scene} is a large scale dataset containing 150 classes of both \emph{things} and \emph{stuff} (uncountable or amorphous regions like sky or grass). Cityscapes~\cite{cordts2016cityscapes} has 19 classes having both \emph{things} and \emph{stuff} and covering scenes from 21 different urban cities.

\noindent\textbf{CISS Protocols:} Two different CISS settings introduced by~\cite{cermelli2020modeling} are \emph{disjoint} and \emph{overlapped}. While the \emph{disjoint} setting assumes that the future classes are known and removes images with future classes from the current step, the \emph{overlapped} setting is more realistic and has no such assumption. Similar to~\cite{douillard2021plop, cha2021ssul}, we also follow the \emph{overlapped} setting in our experiments. We denote the different settings as X-Y where X is the number of classes in the initial step followed by Y number of classes at every step until all the classes are seen. We train 15-5 (15 classes followed by 5 classes), 15-1 (15 then 1 class in each step), 5-3 and 10-1 settings on VOC. Similarly, we train 100-50, 100-10, 100-5 and 50-50 on ADE20K and 14-1 and 10-1 settings on Cityscapes.

\noindent\textbf{Metrics:}
The mean Intersection over Union (mIoU) metric is calculated after the last step for the initial set of classes, the incremental classes, and for all the classes. The mIoU for the initial classes reflects the stability of model to catastrophic forgetting. The mIoU for the incremental classes reflects the plasticity of the model to learn new classes and the overall mIoU metric signifies the overall performance.

\noindent\textbf{Implementation Details:}
Deeplab-v3~\cite{chen2017rethinking} with a ResNet-101~\cite{he2016deep} backbone pretrained on ImageNet~\cite{deng2009imagenet} having output stride of 16 is used for the experiments. Similar to~\cite{zhang2022representation}, we use a higher initial learning rate and obtain an improved baseline for MiB. We train MiB and MiB+AWT models with SGD and a learning rate of $2 \times 10^{-2}$ for the first step only and $10^{-3}$ for the incremental steps. The models are trained with a batch size of 24 using 2 GPUs for 30 epochs per step for VOC and Cityscapes and 60 epochs for ADE20K. Specific to SSUL models, we follow the same training settings as~\cite{cha2021ssul} since it performs freezing of weights and requires different training hyperparameters. The final results are reported on the validation set of the datasets. Since Cityscapes does not have a real background class, we merge the unlabeled classes into a virtual background class.

We use layer integrated gradients from~\cite{kokhlikyan2020captum} for obtaining the attributions and a threshold $k$ to select the 25\% most significant channels for new classes based on experiments provided in the supplementary material. We obtain a unique set of channels mask for each new class for all settings having 5 or lesser class increments. For settings like 100-10, 100-50 and 50-50 on ADE20K, we use a single channel mask for all the new classes. Code is publicly available\footnote{\url{https://github.com/dfki-av/AWT-for-CISS}}.

\noindent\textbf{Baselines:} We compare our approach with the recent state-of-the-art methods ILT~\cite{michieli2019incremental}, MiB~\cite{cermelli2020modeling}, SDR~\cite{michieli2021continual}, PLOP~\cite{douillard2021plop}, SSUL~\cite{cha2021ssul}, RCIL~\cite{zhang2022representation} and UCD~\cite{yang2022uncertainty}. We apply AWT on two methods, MiB~\cite{cermelli2020modeling} and SSUL~\cite{cha2021ssul}.
We also compare with the upper bound (Joint model learned in non-incremental manner). We do not consider approaches using data from past steps~\cite{maracani2021recall} or auxiliary unlabeled data~\cite{yu2022self}.

\ade{Experimental results on ADE20K. Improvements using AWT \underline{underlined}. Best among columns in \textbf{bold}. $\dagger$:~results excerpted from~\cite{zhang2022representation}. * implies results come from re-implementation. Other results come from the respective papers.}

\subsection{Quantitative Evaluation}

\noindent\textbf{Pascal VOC 2012:} We show the quantitative experiments on VOC 15-5, 15-1, 5-3 and 10-1 settings in \cref{tab:voc}. We observe that while ILT struggle on all settings, other methods show significant improvements. Pooling-based distillation methods like PLOP and RCIL do better in 15-5, 15-1 and 10-1 settings but these methods perform poorly on the 5-3 setting where the number of classes is less in the initial step.

AWT with MiB outperforms MiB significantly on all the settings. On 15-5, our model outperforms MiB by 1.5 percentage point ($p.p$) on the overall \miou{} metric. On the 15-1 setting, our model reduces the forgetting of the initial classes by 11 $p.p$ while the overall performance improves by 8.7 $p.p$. On the 5-3 setting having multiple class increments, AWT improves the overall \miou{} by 4.3 $p.p$ over MiB. On the most challenging setting of 10-1 having 11 steps, AWT reduces the forgetting of the initial classes by 19.1 $p.p$ and improves the learning of new classes by 4.2 $p.p$. 

AWT with SSUL~\cite{cha2021ssul} performs similar to SSUL for the 15-5, 15-1 and 5-3 settings, while for the challenging 10-1 setting, it reduces forgetting of the initial classes by 1.8 $p.p$ and improves the performance on new classes by 1.0 $p.p$. SSUL makes use of saliency maps targeted for \emph{things} or objects and moves them from background to an unknown class for representing the future classes. This label augmentation improves performance on all settings of VOC since this dataset has only object classes. On the contrary, this saliency-based modelling is not applicable for ADE20K, Cityscapes and other datasets which have both \emph{things} and \emph{stuff} classes, and SSUL suffers from high forgetting as we observe in~\cref{tab:ade20k,tab:cityscapes}.

\noindent\textbf{ADE20K:} ADE20K~\cite{zhou2017scene} is a difficult dataset with 150 classes and has the joint model \miou{} of only 38.9\%. We report the experimental results on ADE20K 100-50, 100-10 and 50-50 in \cref{tab:ade20k} with analysis of performance on the incremental sets of classes. We also consider a long setting of 100-5 (11 tasks) in \cref{tab:ade20k2}.

\secondade{Experimental results on the 100-5 setting on ADE20K. Improvements using AWT \underline{underlined}. Best among columns in \textbf{bold}. $\dagger$:~results excerpted from~\cite{zhang2022representation}. * implies results come from re-implementation.}

On 100-50, our model improves the overall performance over MiB by 0.3 $p.p$.
On 50-50 setting, our model achieves an overall improvement of 1.7 $p.p$ over MiB and 1.0 $p.p$ over RCIL. Moving to the longer sequence of 100-10 with 6 steps, our model improves MiB by 3.6 $p.p$ and PLOP+UCD by 0.9 $p.p$. On the 11 step setting of 100-5, AWT improves MiB by 4.6 $p.p$ and its nearest contender SSUL by 1.0 $p.p$. MiB+AWT achieves state-of-the-art results on all settings of ADE20K indicating the robustness towards predicting both \emph{things} and \emph{stuff} classes.

\noindent\textbf{Cityscapes:} We perform CISS experiments on two long sequence settings of 14-1 (6 tasks) and 10-1 (10 tasks) of Cityscapes~\cite{cordts2016cityscapes} dataset. We introduce the 10-1 setting where we initially train on 10 classes (road, sidewalk, building, wall, fence, pole, light, sign, vegetation, terrain) and add each of the 9 classes (sky, person, rider, car, truck, bus, train, motorcycle, bicycle) one at a time. We evaluate naive fine-tuning (FT), PLOP, RCIL, SSUL, MiB, and AWT with both SSUL and MiB, and report the \miou{} results in \cref{tab:cityscapes}.

We observe that while FT has very low overall \miou{}, PLOP, RCIL and MiB have improved overall performance on both settings. SSUL shows higher performance on incremental classes but with very high forgetting on the initial classes compared to others.
On the 14-1 setting, AWT with SSUL improves the overall \miou{} over SSUL by 0.8 $p.p$ and MiB+AWT outperforms MiB by 1.5 $p.p$ with a significant improvement of 7.3 $p.p$ on the performance of incremental classes (15-19). On the longer 10-1 setting, SSUL+AWT increases the overall \miou{} by 0.5 $p.p$ over SSUL while MiB+AWT improves over MiB by 3.0 $p.p$ with a good margin of 7.1 $p.p$ improvement on the incremental classes (11-19). AWT significantly improves the plasticity of the models to better learn the new classes in both settings.

\cityscapes{Experimental results on Cityscapes. Improvements using AWT \underline{underlined}. Best among columns in \textbf{bold}. All results come from our implementation.}

\subsection{Ablation Study}\label{sec:ablation}
We analyze the effectiveness of our approach with ablation experiments on Pascal-VOC 2012 for the 15-1 setting. 

\noindent\textbf{Selective weight transfer:} We analyze the importance of the selective weight transfer approach in \cref{tab:addablate}. 
The weight transfer proposed by MiB~\cite{cermelli2020modeling} is a better choice compared to the case when no weights are transferred. We show that our proposed AWT ensures the selection of the most significant channels for new classes by performing experiments with random selection of channels without using attributions. We observe that randomly selecting the same number of channels (25\% of total channels) and transferring their weights in the same way as AWT performs poorly on both initial and incremental sets of classes.

\noindent\textbf{Design choices:} We consider the alternative ways of selecting the significant channels and analyze them in \cref{tab:ablate}. In AWT, we take the mean of the attribution maps from all images of the current step and then perform max-pooling. Here, we consider the alternative of pooling the channels first for all the images and then take the mean of the pooled values. We also consider using average-pooling instead of max-pooling. We experimentally show that mean followed by max-pool is the best choice for channel selection.

\addablate{Ablation study for selective channel weights transfer on Pascal-VOC 2012.}

\ablate{Ablation study for different design choices using MiB~\cite{cermelli2020modeling} + AWT on Pascal-VOC 2012.}

\begin{figure}[t]
\centering
\includegraphics[width=0.95\columnwidth]{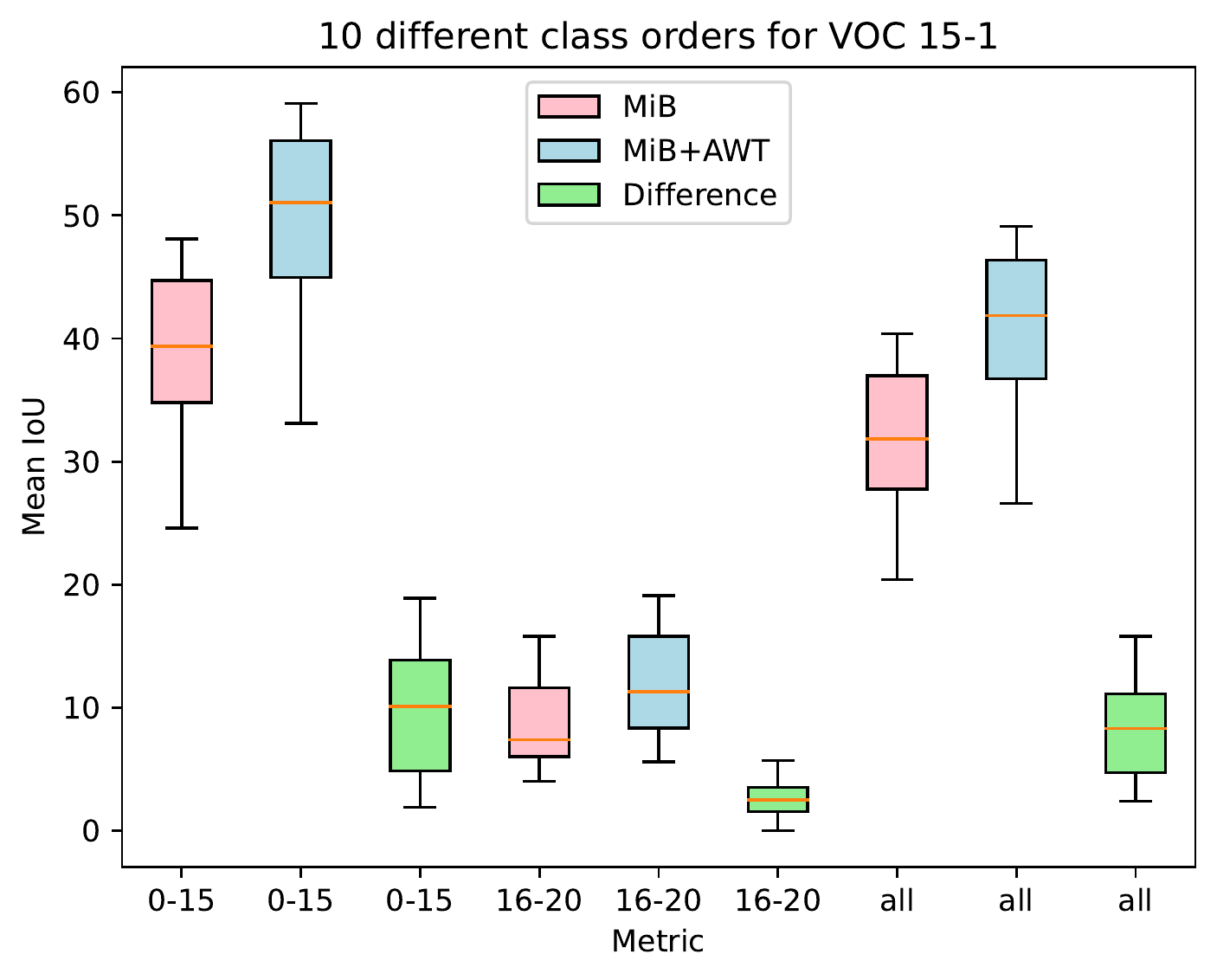}
\caption{Boxplots of the \miou{} of initial, new, and all classes for 10 random class orders.}
\label{fig:boxplot}
\end{figure}

\begin{figure*}[t]
    \includegraphics[width=\textwidth]{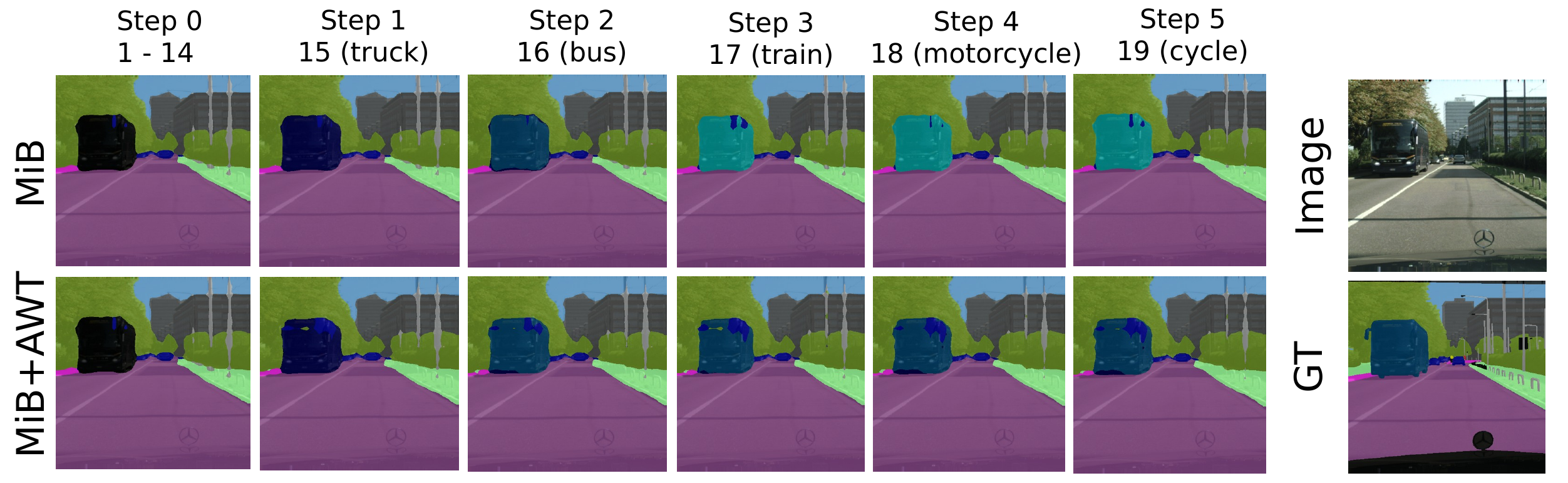}
    \caption{Visualization of predictions using MiB and MiB+AWT in 14-1 setting for Cityscapes. MiB is highly biased towards the new classes and classifies the \emph{bus} as \emph{train} (row 1) while MiB+AWT correctly classifies the \emph{bus} (row 2).}
    \label{fig:combo2}
\end{figure*}

\noindent\textbf{Random class ordering:} The order of classes plays an important role in CISS settings. We experiment with 10 different class orderings on VOC 15-1 setting. We show the average performance for MiB and MiB+AWT in \cref{fig:boxplot}. We also plot the difference between MiB+AWT and MiB for every class order to demonstrate the robustness of our method using random class sequences.

\noindent\textbf{Computational Complexity:} The time taken for the attribution module depends on the number of new-class images. We use two Nvidia RTXA6000 GPUs for training the models. For each image, it takes approximately 0.68 seconds to compute the attributions.
For VOC 15-1, MiB+AWT takes 10.32 hours for training while the attribution module for all steps only takes 37 minutes (6\% of the total training time). Thus, the computational time for the attribution module is considerably less compared to the entire training process.

For further analysis, refer to the supplementary paper.

\subsection{Qualitative Evaluation}
\cref{fig:combo2} shows the predictions of MiB and MiB+AWT across time on Cityscapes 14-1 setting. MiB is biased towards the new classes and forgets the class (\emph{bus}) learned in step 2 and classifies the \emph{bus} as \emph{train} from step 3 onward. MiB+AWT still classifies the \emph{bus} correctly till step 5.

\cref{fig:ade_combo} shows the predictions for both MiB and MiB+AWT models trained in 100-5 setting on test images of ADE20K. We show that MiB+AWT improves predictions of classes like \emph{fan} (row 1), \emph{wardrobe} (row 2) and \emph{chair, chandelier} (row 3) compared to MiB.

\adecombo{}

\section{Conclusion and Limitations}
In this paper, we addressed the issue of semantic background shift during the initialization of the new foreground classifiers at each step of CISS. We discussed the problems with the existing initialization method, and then proposed an attribution-based approach to identify weights that are of interest for the new foreground classes and transfer these weights. This selective initialization takes into account the gradual transition of classes from background to foreground across time. Experimental results on multiple datasets showed that our approach reduces the forgetting of old classes and further improves the plasticity. Our weight transfer approach generalizes well with both \emph{things} and \emph{stuff} classes and outperforms the existing CISS methods. This work lays the foundation for attribution-aware weight initialization that could be more generally used for incremental learning problems involving multi-class classification

Although our method works well with most incremental settings, the strategy of selecting multiple set of channels for multiple new classes would involve a significant increase in computational complexity as a function of the number of new classes at every step, especially for 10, 50 or more class increments at a step. We believe future work can be done to address this limitation. We hope that our attribution-based channel selection approach would be adapted beyond semantic segmentation to other computer vision applications.

\minisection{Acknowledgement.}
{\footnotesize This work was partially funded by the Federal Ministry of Education and Research Germany under the project DECODE (01IW21001) and partially by the Spanish Government funded project PID2019-104174GB-I00/AEI/10.13039/501100011033.}

\ifpreprint
    \newpage
\twocolumn[
    \centering
    \Large
    \bf
    Supplementary Materials
    \vskip .5em
    \vspace*{12pt}
]

\setcounter{section}{0}
\renewcommand\thesection{\Alph{section}}

\section*{Introduction}
In this supplementary material to our main paper \textit{Attribution-aware Weight Transfer: A Warm-Start Initialization for Class-Incremental Semantic Segmentation}, we discuss the details of the gradient-based attribution method, Integrated Gradients~\cite{sundararajan2017axiomatic} used in our Attribution-aware Weight Transfer (AWT) initialization.
We further share more details of our implementation for better reproducibility, and perform additional ablative experiments to analyze the impact of the proposed warm-start initialization.
Finally, we present the qualitative results of AWT with MiB~\cite{cermelli2020modeling} and SSUL~\cite{cha2021ssul} on Pascal-VOC 2012.

\section{Attribution Method}
\noindent\textbf{Integrated Gradients:}
Consider a deep neural network as a function $F: \reals^n \rightarrow [0,1]$. Let $x \in \reals^n$ be the input image and $\xbase \in \reals^n$ be a baseline black image. Integrated Gradients (IG)~\cite{sundararajan2017axiomatic} computes and accumulates the gradients at all points along the straight line path (in $\reals^n$) from the baseline to the input.

Let $\tfrac{\partial F(x)}{\partial x_i}$ be the gradient of $F(x)$ along the $i^{th}$ dimension. Then the integrated gradient along the $i^{th}$ dimension for an input $x$ and baseline $\xbase$ is defined as follows:
\begin{equation}
\small
IG_i(x) \synteq (x_i-\xbase_i)\times\int_{\sparam=0}^{1} \tfrac{\partial F(\xbase + \sparam\times(x-\xbase))}{\partial x_i  }~d\sparam
\end{equation}
Note that the attributions add up to the difference between $F(x)$ and $F(\xbase)$.

\noindent\textbf{Layer Integrated Gradients:} Layer Integrated Gradients~\cite{kokhlikyan2020captum} is designed for computing attributions corresponding to inputs or outputs of a specific layer of the network. For a given layer, the size of the attribution maps is the same as the layer's input or output dimensions, based on whether we attribute to the inputs or outputs of that layer. In our method, we compute the attributions for the inputs to the final classifier layer. We obtain the attributions corresponding to a given target class (\emph{background} class in our method).

\section{Reproducibility}
\noindent\textbf{Datasets:} We evaluate our models on Pascal-VOC 2012~\cite{everingham2015pascal}, ADE20K~\cite{zhou2017scene} and Cityscapes~\cite{cordts2016cityscapes}. VOC contains 10,582 images for training and 1,449 images for testing. ADE20K contains 20,210 and 2,000 images for training and testing respectively. Cityscapes contains 2,975 training images and 500 testing images.

\noindent\textbf{Implementation details:} We use Deeplab-v3~\cite{chen2017rethinking} with ResNet-101~\cite{he2016deep} backbone pretrained on ImageNet~\cite{deng2009imagenet} having output stride of 16.
In-place activated batch normalization~\cite{bulo2018place} is used to reduce memory requirements.
Similar to~\cite{cermelli2020modeling, douillard2021plop, zhang2022representation}, the data augmentation from~\cite{chen2017rethinking} has been applied along with random cropping of $512 \times 512$ pixels for training and a center crop of the same size for testing.
A random horizontal flip is performed on the training set only.

We re-implement SSUL by training for 60 epochs on ADE20K dataset. We follow the same training settings for SSUL as proposed in~\cite{cha2021ssul} for VOC and ADE20K. For Cityscapes, we trained SSUL with a learning rate of 0.01 and a batch size of 24. We train the other models of FT, PLOP, RCIL for Cityscapes with SGD and a learning rate of $2 \times 10^{-2}$ for the first step only and $10^{-3}$ for the incremental steps.

\noindent\textbf{Class order:} For all the quantitative experiments, we order the classes by increasing class id, \ie the default order of the respective datasets.

For the ablation experiment using random orders on VOC 15-1, we sampled the following 10 class sequences: \\
{\small[1, 2, 3, 4, 5, 6, 7, 8, 9, 10, 11, 12, 13, 14, 15, 16, 17, 18, 19, 20]}\\
{\small[12, 9, 20, 7, 15, 8, 14, 16, 5, 19, 4, 1, 13, 2, 11, 17, 3, 6, 18, 10]}\\
{\small[13, 19, 15, 17, 9, 8, 5, 20, 4, 3, 10, 11, 18, 16, 7, 12, 14, 6, 1, 2]}\\
{\small[15, 3, 2, 12, 14, 18, 20, 16, 11, 1, 19, 8, 10, 7, 17, 6, 5, 13, 9, 4]}\\
{\small[7, 13, 5, 11, 9, 2, 15, 12, 14, 3, 20, 1, 16, 4, 18, 8, 6, 10, 19, 17]}\\
{\small[7, 5, 9, 1, 15, 18, 14, 3, 20, 10, 4, 19, 11, 17, 16, 12, 8, 6, 2, 13]}\\
{\small[12, 9, 19, 6, 4, 10, 5, 18, 14, 15, 16, 3, 8, 7, 11, 13, 2, 20, 17, 1]}\\
{\small[13, 10, 15, 8, 7, 19, 4, 3, 16, 12, 14, 11, 5, 20, 6, 2, 18, 9, 17, 1]}\\
{\small[1, 14, 9, 5, 2, 15, 8, 20, 6, 16, 18, 7, 11, 10, 19, 3, 4, 17, 12, 13]}\\
{\small[16, 13, 1, 11, 12, 18, 6, 14, 5, 3, 7, 9, 20, 19, 15, 4, 2, 10, 8, 17]}\\ 

\section{Additional Ablation Experiments}
Additional experiments are performed to analyze the effect of the initialization and the number of training iterations per step.
We show in \cref{tab:initablate} that training the model with random initialization for a higher number of iterations ($\times 2$, $\times 4$) cannot reach the performance of AWT initialization or even the one proposed by~\cite{cermelli2020modeling}.
Instead, training for more iterations causes higher forgetting of old classes. 

Furthermore, we vary the threshold $k$ to select the most significant 10\%, 25\%, 50\% and 75\% of the channels for weight transfer.
Based on the results of this experiment shown in \cref{tab:thresholdablate}, our final AWT uses a ratio of 25\% for all our experiments in the main paper.

\setcounter{table}{6} 

\initablate{Ablation study for significance of weight transfer on Pascal-VOC 2012.}

\thresholdablate{Ablation study for selection of threshold using MiB+AWT on Pascal-VOC 2012.}

\setcounter{figure}{6}

To discuss the role of AWT on reducing the effect of background shift, we analyze the performance of the newly added classes after every step of training for VOC 15-1 and ADE20K 100-10 settings in \cref{fig:analysis}. We observe that MiB+AWT better learns the new set of classes which transitions from the previous background to current foreground. This indicates reduced effect of the background shift with AWT across multiple steps.

\begin{figure}[t]
\centering
    \begin{subfigure}{\columnwidth}
        \centering
        \includegraphics[width=\columnwidth]{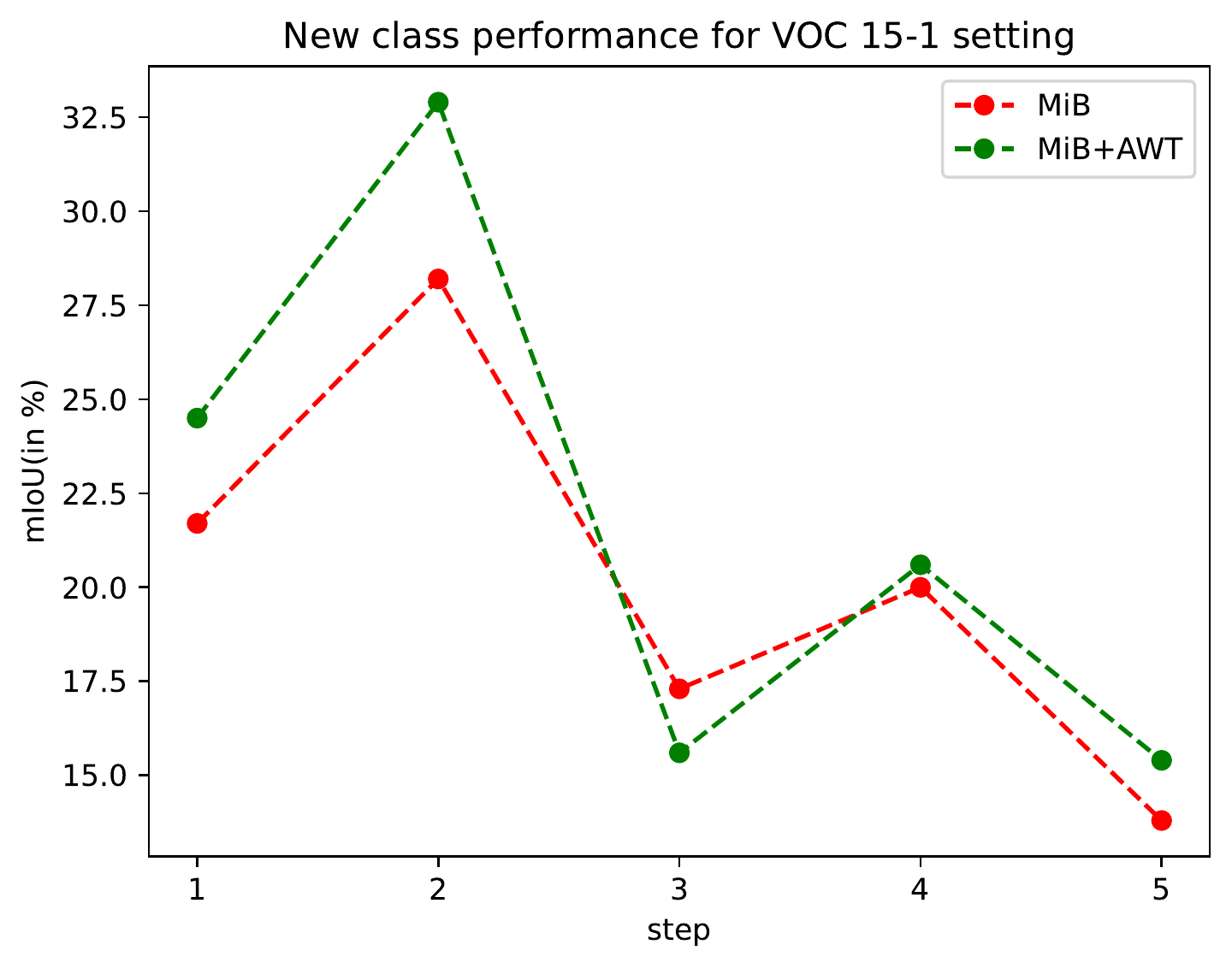}
        \subcaption{VOC 15-1 setting}
        \label{fig:voc}
    \end{subfigure}
    \hfill
    \begin{subfigure}{\columnwidth}
        \centering
        \includegraphics[width=\columnwidth]{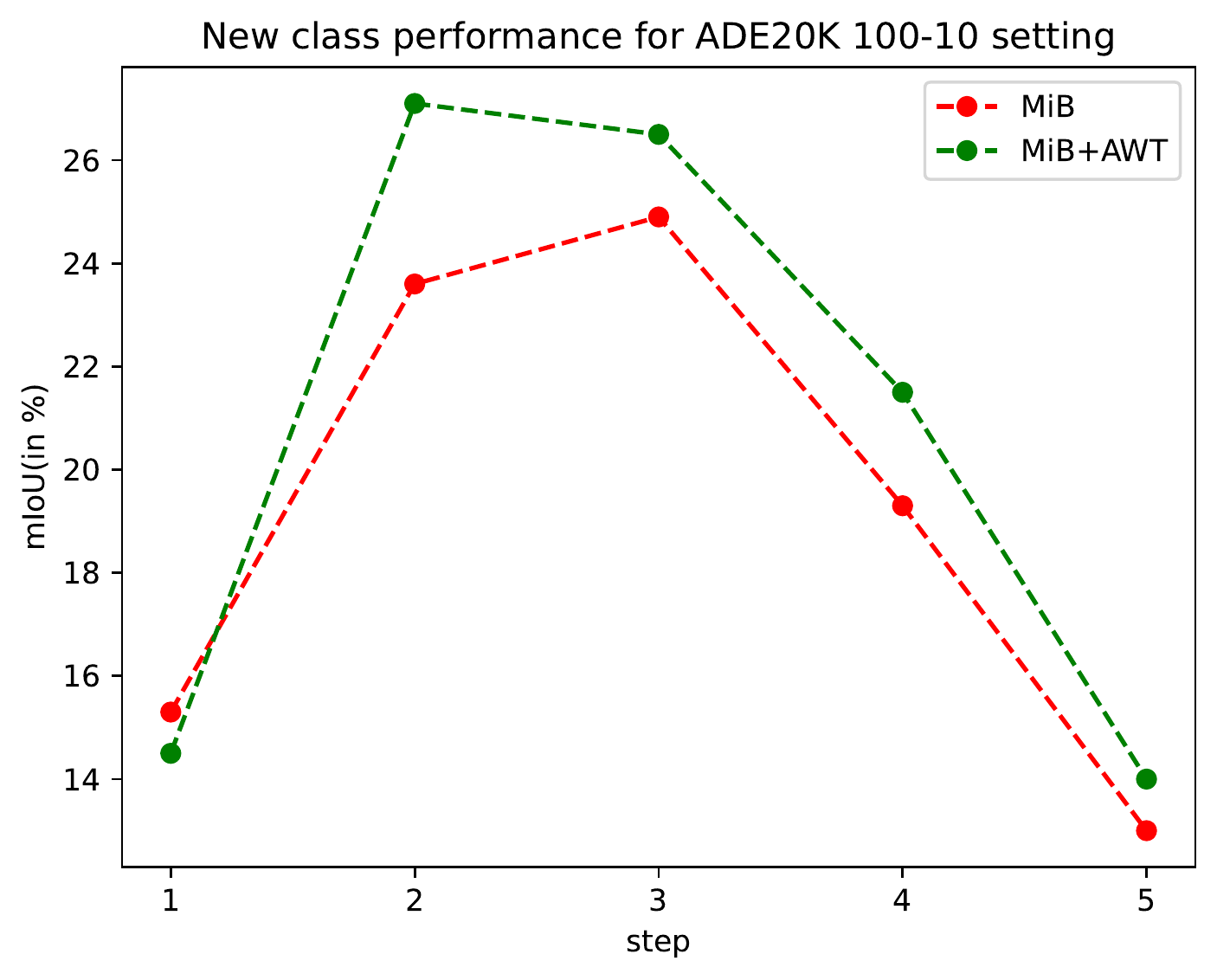}
        \subcaption{ADE20K 100-10 setting}
        \label{fig:ade}
    \end{subfigure}
\caption{Analysis of the learning of new classes at every step}
\label{fig:analysis}
\end{figure}

\section{Additional Qualitative Evaluation}
\Cref{fig:voc_combo} shows the comparison of predictions using MiB, MiB+AWT, SSUL, and SSUL+AWT on some test samples of Pascal-VOC 2012 using models trained in the 10-1 setting.
Over both the methods, AWT improves the predictions for multiple classes like \emph{TV}, \emph{car}, \emph{aeroplane}, \emph{bird}, \emph{chair}, \emph{table}, \emph{horse}, \emph{person}, \emph{dog}, and many more.

\voccombo{}
    {\small
    \bibliographystyle{ieee_fullname}
    \bibliography{egbib_supp}
    }
\else
    {\small
    \bibliographystyle{ieee_fullname}
    \bibliography{egbib}
    }
\fi

\end{document}